\documentclass{article} 
\usepackage{iclr2027_conference,times}
\iclrfinalcopy

\usepackage{amsmath,amsfonts,bm}

\def\eqref#1{equation~\ref{#1}}

\def\1{\bm{1}}

\DeclareMathAlphabet{\mathsfit}{\encodingdefault}{\sfdefault}{m}{sl}
\SetMathAlphabet{\mathsfit}{bold}{\encodingdefault}{\sfdefault}{bx}{n}

\DeclareMathOperator*{\argmax}{arg\,max}
\DeclareMathOperator*{\argmin}{arg\,min}

\renewcommand{\eqref}[1]{Eq.~(\ref{#1})}

\usepackage{hyperref}
\usepackage{url}
\usepackage[fixed]{fontawesome5}
\usepackage{graphicx}
\usepackage[normalem]{ulem}
\usepackage{algorithm}
\usepackage{algpseudocode}
\usepackage{makecell, multirow, booktabs}
\usepackage[table,dvipsnames]{xcolor}
\newcommand{\rev}[1]{#1}
\usepackage{arydshln}
\usepackage{array}
\usepackage{caption}

\newcommand{\fourblock}[2]{
    \begin{tabular}{@{}c@{}}
        \small\texttt{#2} \\[4pt]
        \includegraphics[width=\dimexpr 0.3\textwidth+2pt\relax]{#1}
    \end{tabular}%
}

\title{Chameleon: Dynamic Format Adapter for\\Efficient Diffusion}

\author{Arnab Sanyal \& Sandeep Chinchali\\
	UT SWARM Lab, Department of Electrical \& Computer Engineering \\
	The University of Texas, Austin TX 78712\\
	\texttt{\{sanyal, sandeepc\}@utexas.edu}
}

\begin{document}

\maketitle

\begin{figure}[htbp]
  \centering
  \begin{minipage}[t]{0.15\textwidth}
    \centering
    \includegraphics[width=\textwidth, height=\textwidth]{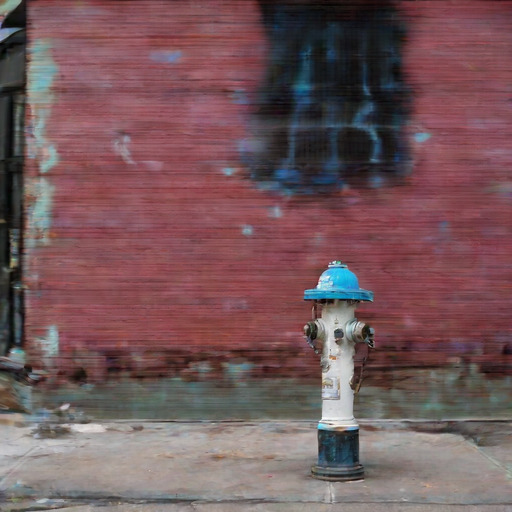}
    \par\smallskip\small \texttt{Q-Diffusion}
  \end{minipage}%
  \hspace{2pt}
  \begin{minipage}[t]{0.15\textwidth}
    \centering
    \includegraphics[width=\textwidth, height=\textwidth]{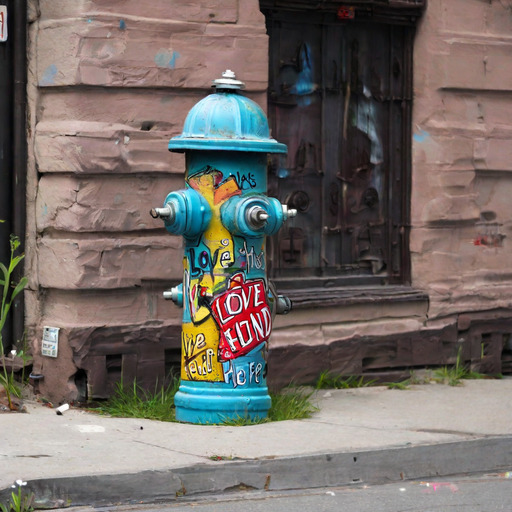}
    \par\smallskip\small Ours
  \end{minipage}%
  \hspace{12pt}
  %
  \begin{minipage}[t]{0.15\textwidth}
    \centering
    \includegraphics[width=\textwidth, height=\textwidth]{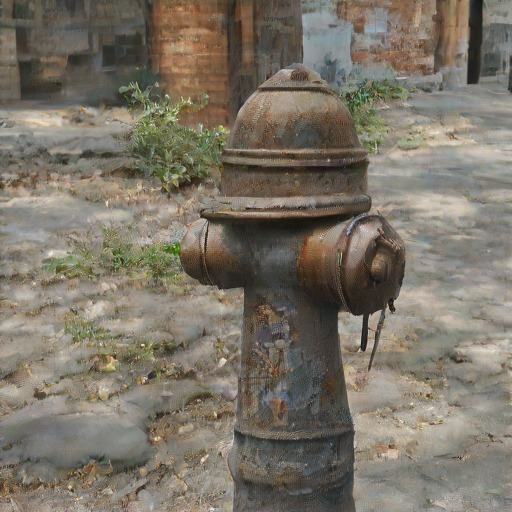}
    \par\smallskip\small \texttt{MixDQ}
  \end{minipage}%
  \hspace{2pt}
  \begin{minipage}[t]{0.15\textwidth}
    \centering
    \includegraphics[width=\textwidth, height=\textwidth]{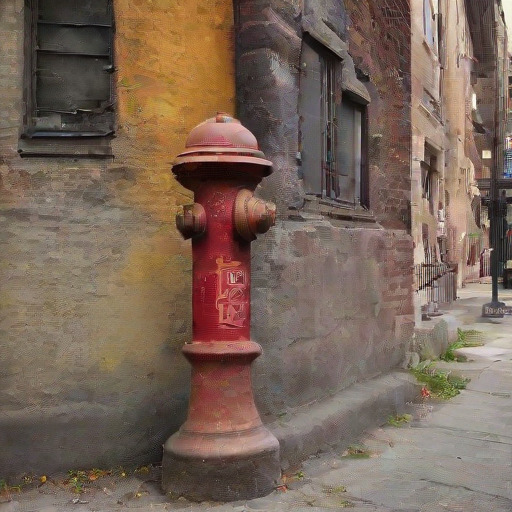}
    \par\smallskip\small Ours
  \end{minipage}%
  \hspace{12pt}
  %
  \begin{minipage}[t]{0.15\textwidth}
    \centering
    \includegraphics[width=\textwidth, height=\textwidth]{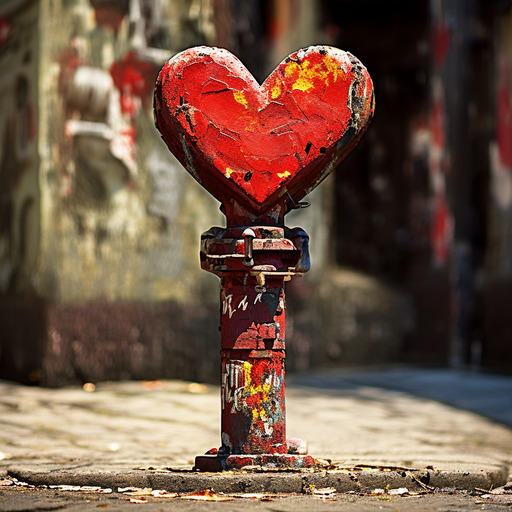}
    \par\smallskip\small \texttt{Q-DiT}
  \end{minipage}%
  \hspace{2pt}
  \begin{minipage}[t]{0.15\textwidth}
    \centering
    \includegraphics[width=\textwidth, height=\textwidth]{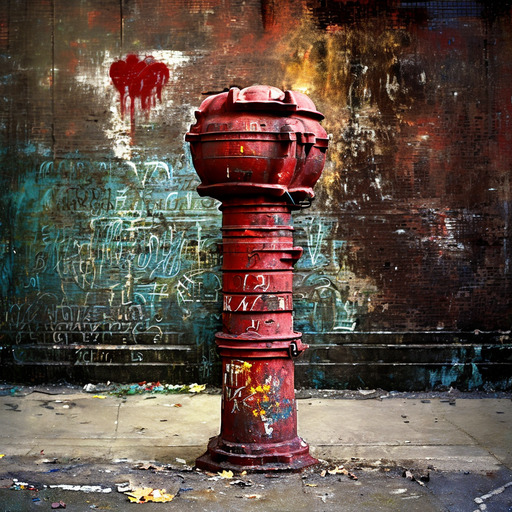}
    \par\smallskip\small Ours
  \end{minipage}

  \caption{Images generated by each per-architecture \texttt{PTQ} baseline and by Chameleon at 4-bit weights (W$_{4}$A$_{8}$), from the COCO-2014 caption ``An old city fire hydrant has `Love was found here' painted on it."}
  \label{fig:intro-image}
\end{figure}

\begin{abstract}
A very popular way to enable image generation on memory-constrained accelerators using modern spatiotemporal generative models is post-training quantization (\texttt{PTQ}). However, every existing diffusion \texttt{PTQ} scheme shares an unstated commitment: the \textit{number format} is fixed in advance, and only the scale, zero point, or the per-layer bit-width is allowed to move. This is unfortunate, because at a fixed bit-width the best format depends on the distribution being encoded, and that distribution differs from tensor to tensor: across weight channels, across layers, and -- in a denoising model -- along the \textit{diffusion timestep}, where activations slide from heavy-tailed and noise-dominated at $t\rightarrow T$ to tightly clustered and structured at $t\rightarrow 0$. One format for the whole network is therefore mispriced for most of its tensors, and re-tuning a scale cannot \rev{fully} repair a format mismatch. In this work, we propose a \texttt{PTQ} framework, which we call \textbf{Chameleon}, that holds the bit-width fixed and treats the number format itself as a discrete variable, chosen per weight channel and \rev{per activation layer (optionally per timestep bucket)}. Chameleon draws activation formats from a palette of \{\texttt{INT8}, \texttt{FP8} \texttt{E4M3}, \texttt{FP8} \texttt{E5M2}, \texttt{MXFP8}, \texttt{MXINT8}\} and weight formats from \{\texttt{INT8}, \texttt{MXINT8}\} at $8$ bits or \{\texttt{INT4}, \texttt{NF4}, \texttt{FP4 E2M1}, \texttt{MXINT4}, \texttt{MXFP4}\} at $4$ bits, choosing activation formats ahead of time from two cheap statistics -- empirical activation kurtosis and closed-form diffusion SNR $\bar\alpha_t/(1-\bar\alpha_t)$ -- stored in a per-(layer, timestep bucket) lookup table, and weight formats offline by reconstruction error. An architectural fork adapts the same selection layer to each model family, so each model exercises the palette as far as its own statistics warrant: weights and per-(layer, bucket) activations on multi-step \texttt{SDXL}, weights alone where a single sampler step leaves no timestep axis. \rev{Most of the gain is per-layer; timestep refinement helps mainly at 6-bit activations.} We evaluate Chameleon across three diffusion regimes -- multi-step \texttt{SDXL}, single-step \texttt{SDXL-Turbo}, and the \texttt{PixArt}-$\mathtt{\alpha}$ Diffusion Transformer -- on COCO-2014.  Chameleon achieves the best \texttt{FID} for all six backbone$\times$bit-width settings, while preserving semantic alignment with the text prompts (\texttt{CLIP} within $0.24$ of the \texttt{FP16} reference in every setting, and the best of all quantized methods at  W$_{4}$A$_{8}$). \rev{On \texttt{PixArt}-$\mathtt{\alpha}$, a real kernel is $1.09\times$ faster than \texttt{FP16}.} Adapting the format, rather than only its scale, recovers distributional fidelity at 4-bit weights while holding prompt adherence at the \texttt{FP16} level.\footnote{\href{https://arnabsanyal.com/preprint/chameleon.html}{\faGlobe \texttt{website}}\quad\href{https://github.com/arnabsanyal/Chameleon}{\faGithub \texttt{code}}}
\end{abstract}

\section{Introduction}

The frontier of generative AI lies not in the cloud but on the edge \citep{oasis}. Latent diffusion models, few-step distilled models, and diffusion transformers (\texttt{DiT}s) all generate images from pure noise by stepping through a learned vector field for tens to hundreds of forward passes through a network holding billions of parameters: \texttt{SDXL} alone has roughly $2.6$ billion, and text encoding and the variational auto-encoder add another billion. Replacing 16-bit numerics with 8-, 6-, or 4-bit ones -- post-training quantization (\texttt{PTQ}) -- is the standard response, cutting memory footprint and latency while raising throughput \citep{nagel2021white, frantar2023gptq, xiao2023smoothquant}. This paper proposes a \texttt{PTQ} scheme for the iteratively denoising generators in that list. The key idea is that while weights can be statically quantized, activations should not be, because of the iterative nature of the inference pipeline. We call our \textbf{dynamic format adapter} framework \uline{Chameleon}: at a fixed bit-width it chooses a number format per tensor -- per weight channel, per layer, and per timestep bucket where a timestep axis exists -- using cheaply computed statistical heuristics.

\begin{figure}[t!]
  \centering
  \includegraphics[width=\textwidth]{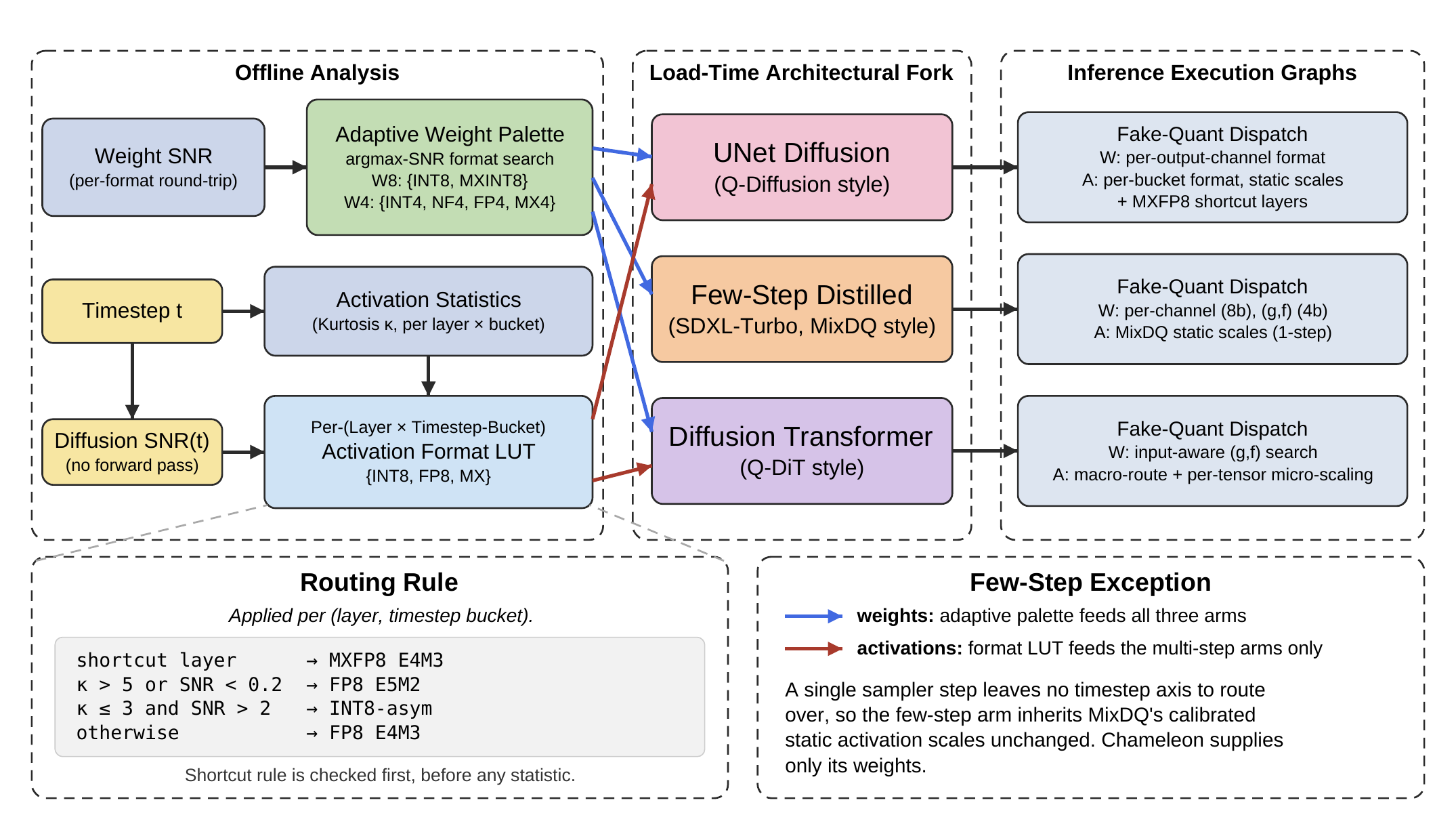}
  \caption{\textbf{Chameleon framework:} Offline, a one-pass calibrator accumulates per-(layer, bucket) activation moments for the empirical kurtosis $\kappa_l(t)$, while the diffusion $\mathrm{SNR}(t)$ is read from the noise schedule. The router applies the rule of Section~\ref{subsec:rthresh} to both and writes the format into an integer lookup table $F \in \{0,\ldots,4\}^{L \times B}$ (code 4 is unused); weights are selected offline (Sections~\ref{subsec:weightSel} and~\ref{subsec:archFork}). At inference, one fake-quantization module reads the \texttt{LUT} per layer per step on the multi-step paths and dispatches to the matching routine.}
  \label{fig:schematic}
\end{figure}

Prior work has contributed a great deal to this space (Section~\ref{sec:related}), but the four diffusion \texttt{PTQ} baselines we compare against -- \texttt{Q-Diffusion}, \texttt{PTQ4DM}, \texttt{MixDQ} and \texttt{Q-DiT} -- all take quantization to mean rounding to the nearest \texttt{INT8} format. This is suboptimal for at least two reasons.

First, at a fixed bit-width the format is still free, and the right choice follows the distribution being encoded: \citet{Nagel2022powerOfExponent} show that the signal-to-quantization-noise ratio (\texttt{SQNR}) at a given width is a function of both that distribution and the format, and \citet{zhang2023mofq} turn this into a per-layer \texttt{INT}-or-\texttt{FP} choice for \texttt{LLM}s. Tensors within one network differ in shape -- across weight channels and across layers -- so no single format is right for all of them. Denoising networks add a second axis: the activations of a single layer are themselves a parametric family of distributions indexed by timestep $t$ \citep{li2023qdiffusion}. At $t \to T$, activation tensors are dominated by Gaussian noise, and the internal activations they induce are heavy-tailed, with kurtosis ($\kappa$) above 5; at $t \to 0$ they are dominated by the structured signal of a near-convergence latent, with $\kappa$ often below 3. \rev{A format that is right for one layer or phase is wrong for another.} Re-tuning the scale of \texttt{INT8} cannot make it tolerate heavy tails, any more than re-scaling \texttt{FP8} \texttt{E5M2} can give its logarithmic bins the density to render fine texture. The fix is to \emph{change formats} \rev{per tensor and, where the statistics warrant, along the schedule}.

Second, microscaling formats -- \texttt{MXFP8, MXINT8, MXINT4} \citep{ocp2023mx, rouhani2023mx} -- are now hardware features on Blackwell-class accelerators, with \texttt{FP8} already native on Hopper and Ada. One shared 8-bit exponent per 32 elements lets a single tensor hold values spanning very different magnitudes without one global scale covering them all. This is exactly what the concatenations at every \texttt{UNet} shortcut need, where shallow skip features meet deep upsampled features. Measured on \texttt{SDXL}, the tensor entering a shortcut \texttt{conv1} spans a peak-to-median per-channel range of $58\times$ at the median layer and up to $163\times$ (Appendix~\ref{app:formats}), so one per-tensor scale must cover outliers that most channels never approach. \texttt{Q-Diffusion} split the computational graph for this; with \texttt{MX} formats, one need not.

These two observations -- that the right format depends on the tensor, and on $t$ where a timestep axis exists, and that the format should sometimes be a microscaling block format -- are the substance of this paper. Figure~\ref{fig:schematic} sketches the resulting framework end-to-end.

We present \textbf{Chameleon}, a \texttt{PTQ} framework that treats the number format as a discrete variable, chosen per weight channel and \rev{per activation layer, optionally per timestep bucket,} from the palettes of Section~\ref{subsec:formatp}. Two cheap statistics -- the empirical kurtosis $\kappa_l(t)$ and the closed-form diffusion \texttt{SNR} -- decide it offline, and the result costs one indexed read per layer per step. \rev{Against error-based per-layer selection (\texttt{MoFQ}-style) and timestep-adaptive \texttt{INT8} scales (\texttt{TDQ}-style), this per-layer format rule is what carries the gain (Appendix~\ref{app:rebuttal}).}

\subsection{Motivation}


\texttt{PTQ} research for time-stepped image generators is architecture-specific, but the three lines of work converge on one observation: the activation distributions inside a denoising network are non-stationary along the diffusion trajectory, and that non-stationarity is the dominant source of quantization error. What none of those four adjusts is the number format itself. \texttt{Q-Diffusion}, \texttt{PTQ4DM}, \texttt{MixDQ}, and \texttt{Q-DiT} all commit up-front to a single format -- invariably \texttt{INT8} -- and route their design effort into scales, zero points, per-layer bit-widths, group sizes, and sampler schedules on top of that fixed commitment. This is an odd restriction to accept when the underlying activation family spans regimes as different as heavy-tailed, noise-dominated activations at $t\rightarrow T$ and tightly clustered structured signals at $t\rightarrow 0$. If the family of activation distributions is the problem, the family of number formats is the natural response.


Chameleon respects those architecture-specific lessons rather than discarding them: a hard fork at model load selects a \texttt{UNet}, few-step, or \texttt{DiT} execution path (Section~\ref{subsec:archFork}), and because the format choice sits \emph{underneath} that machinery, each path threads Chameleon's \texttt{LUT} through the treatment inherited from prior work. ``Which format to use" is thereby decoupled from ``which architecture we are quantizing."


This is tractable because the format is a \emph{discrete} variable: the palette is small (five formats for activations, six for weights at $4$ bits) and fixed by what accelerators implement natively, so selection is a routing problem rather than a continuous optimization, and both statistics \rev{that the routing rule reads} are cheap -- kurtosis is a one-pass moment accumulation, and \texttt{SNR} is read off the noise schedule with no forward pass.


\section{Related Work}
\label{sec:related}

Extended versions of the paragraphs below, and a note on \texttt{LLM} quantization, are in Appendix~\ref{app:related}.

\paragraph{Diffusion-model PTQ:} \texttt{Q-Diffusion} \citep{li2023qdiffusion} first recognized that activation distributions inside a denoising \texttt{UNet} are not stationary across the schedule, answering with a uniform-timestep calibration sampler and a split quantizer for the shortcut layers. \texttt{PTQ4DM} \citep{shang2023ptq4dm} added per-bucket calibration statistics. Both take the format itself -- \texttt{INT8} -- as a given and tune scales around it.

\paragraph{Few-step and consistency-model PTQ:} Few-step distilled models -- by consistency \citep{luo2023lcm} or adversarial \citep{sauer2023adversarial} distillation -- are hostile to quantization: each of their one to four steps carries far more of the semantic burden than one step of a 50-step trajectory. \texttt{MixDQ} \citep{zhao2024mixdq} combines metric-decoupled mixed precision, assigned by integer programming, with a \texttt{BOS}-aware bypass in the cross-attention $K$ and $V$ projections; \texttt{Q-Sched} \citep{frumkin2025qsched}, concurrent with and complementary to Chameleon, instead learns a single pair of scalar preconditioning coefficients that warp the sampling trajectory. Our few-step fork composes with \texttt{MixDQ} rather than replacing it (Section~\ref{subsec:archFork}).

\paragraph{DiT quantization:} Diffusion transformers exhibit large variance across the \emph{input} channel dimension of their linear-layer activations, which tensor- or output-channel-wise quantization cannot absorb. \texttt{Q-DiT} \citep{chen2024qdit} introduced \emph{group-wise} weight quantization with an evolutionary search over group sizes $g \in \{32, 64, 128, 192, 288\}$, and \emph{dynamic} on-the-fly scale and zero-point computation for activations -- again with \texttt{INT} as the only number system. \texttt{ViDiT-Q} \citep{zhao2025viditq} reports W$_{8}$A$_{8}$ and W$_{4}$A$_{8}$ on \texttt{PixArt}-$\mathtt{\alpha}$ with finer-grained dynamic quantization, again within \texttt{INT}. Our \texttt{DiT} fork keeps both, generalizing the group search to a joint $(g, f)$ search over group size and format (Section~\ref{subsec:archFork}). \texttt{PTQ4DiT} \citep{wu2024ptq4dit} and \texttt{FP4DiT} \citep{chen2025fp4dit} are not re-run in our harness\rev{; \texttt{ViDiT-Q} is run in its own pipeline with our prompts, seeds and scoring and, because its \texttt{FP16} reference differs, is compared relative to that reference in Appendix~\ref{app:rebuttal}} rather than in Table~\ref{tab:combined_results}.

\paragraph{Microscaling formats:} The \texttt{OCP} \texttt{MX} specification \citep{ocp2023mx, rouhani2023mx} standardized block-shared-exponent narrow formats: each block of 32 elements carries one shared 8-bit exponent, absorbing intra-tensor magnitude variation without per-channel scale infrastructure. To our knowledge, Chameleon is the first \texttt{PTQ} scheme for diffusion models to use \texttt{MX} formats \emph{adaptively}, routing concatenation-bearing \texttt{UNet} shortcut layers to \texttt{MXFP8} by topology.
 
\paragraph{Format selection beyond \texttt{INT}:} At a fixed bit-width the format is itself a free variable: \texttt{MoFQ} \citep{zhang2023mofq} chooses \texttt{INT} or \texttt{FP} per layer for \texttt{LLM}s by quantization error, and \citet{chen2024lowbitfp} search a per-tensor floating-point encoding, layer by layer, for diffusion. \texttt{TDQ} \citep{so2023tdq} comes closest to our temporal axis, but varies the quantization \emph{interval} with $t$, not the format (Appendix~\ref{app:related}).

\paragraph{Where we sit:} Chameleon is the first \texttt{PTQ} scheme for time-stepped image generators to treat the \emph{number format} as a discrete variable indexed by \rev{\emph{layer}, optionally refined per timestep bucket} -- per-layer format selection at a fixed width was introduced for \texttt{LLM}s \citep{zhang2023mofq}, and timestep-dependent quantization \emph{parameters} for diffusion \citep{so2023tdq}, but not the format along the schedule\rev{; against both, re-implemented in our harness, the robustness-based format rule gives lower \texttt{FID} (Appendix~\ref{app:rebuttal})} -- and the first to demonstrate the same routing principle across \texttt{UNet}, few-step \texttt{UNet}, and image \texttt{DiT} under one framework. It is composable rather than competitive: where existing methods care about scales (\texttt{Q-Diffusion}, \texttt{PTQ4DM}), bit-widths (\texttt{MixDQ}), group sizes (\texttt{Q-DiT}), or sampling trajectories (\texttt{Q-Sched}), Chameleon cares about formats, and the underlying calibration loop and group search from each of these methods can sit on top of it without modification.

\section{Background \& Preliminaries}
\label{sec:background}

\subsection{Diffusion Basics}
\label{subsec:diffBasics}
In the variance-preserving discretization used by Stable Diffusion and its derivatives, the noise schedule is summarized by the cumulative product $\bar\alpha_t = \prod_{s \le t}(1 - \beta_s)$, and the \emph{signal-to-noise ratio} admits the closed form
\begin{equation}
\label{eq:snr}
\mathrm{SNR}(t) = \frac{\bar\alpha_t}{1 - \bar\alpha_t},
\end{equation}
which falls monotonically in $t$ across five to eight orders of magnitude along the schedule. We use it as one of our two routing statistics: it is known at calibration time and costs nothing to evaluate. Appendix~\ref{app:diffusion} gives the continuous-time formulation, the per-model \texttt{SNR} ranges, and how \texttt{DiT}s and few-step models fit the same picture.

\subsection{Asymmetry in the Statistics of Weights vs.\ Activations}
\label{subsec:assymWvsAct}

A trained denoising network is a static set of weights $\{W_l\}$, whose statistics do not depend on $t$ and can be characterized once, offline, per channel. Activations are different: for a fixed layer $l$, the input tensor $x_l(t)$ depends on both the random latent and the timestep, and its distribution changes dramatically along the schedule. Appendix~\ref{app:phases} works through an illustrative example on \texttt{SDXL}: measured along the schedule, activation kurtosis and $\mathrm{SNR}(t)$ separate three phases -- heavy-tailed at low \texttt{SNR} (Phase 1), settling at mid \texttt{SNR} (Phase 2), and near-Gaussian at high \texttt{SNR} (Phase 3) -- each best served by a different format, and layers cross them at different times, which is why Chameleon routes per (layer, bucket) rather than per bucket alone.

\section{The Chameleon Framework}
\label{sec:method}
\subsection{Problem Statement}
\label{subsec:pstatment}

Let $\mathbb{N}$ be a denoising network with quantizable layers $l = 1, \ldots, L$. For each layer let $X_l(t) \in \mathbb{R}^{C_1 \times C_2 \times \cdots}$ denote the input activation tensor at timestep $t$ and $W_l \in \mathbb{R}^{C_\mathrm{out} \times C_\mathrm{in} \times \cdots}$ the weight tensor. We discretize the schedule into $B$ uniform-width buckets indexed $b = 0, \ldots, B-1$ and write $b(t)$ for the bucket containing $t$.
 

Given a finite palette $\mathcal{F}$ of number formats -- each with its own scaling granularity $g$ (per-tensor or per-block for activations, per-output-channel or grouped for weights; Section~\ref{subsec:archFork} fixes this per architecture) -- we seek for each (layer $l$, bucket $b$) the format minimizing the mean-squared quantization error
  \begin{equation}
      f^{\star}_{l,b} \;=\; \argmin_{f \in \mathcal{F}}\;
      \mathbb{E}_{t \in b}\!\left[\,
      \bigl\lVert X_l(t) - \mathrm{DQ}_f\!\bigl(\mathrm{Q}_f(X_l(t))\bigr)
      \bigr\rVert_2^2 \,\right],
      \label{eq:format-argmin}
  \end{equation}
where $\mathrm{Q}_f$ and $\mathrm{DQ}_f$ are the quantization and de-quantization operators of format $f$ at its native granularity $g$. The analogous objective is solved per output channel for weights, with the expectation over $t$ dropped because weights are static.
 
Solving \eqref{eq:format-argmin} exhaustively means materializing $X_l(t)$ and computing five round-trips per (layer, bucket). That is feasible for one calibration pass, but two cheap scalars -- the empirical kurtosis $\kappa_l(t)$ of $X_l(t)$ and the diffusion \texttt{SNR} of \eqref{eq:snr}, which depends on $t$ alone -- \rev{stand in for} the argmin \rev{and} populate a static lookup table that ships with the model. \rev{For activations they are not merely a cheaper approximation: selecting formats by calibration error alone gives worse generation quality (Appendix~\ref{app:rebuttal}).}

\subsection{Format Palette}
\label{subsec:formatp}

The activation palette is
  \begin{equation*}
      \mathcal{F}_a \;=\; \{\texttt{INT8-asym},\ \texttt{FP8 E4M3},\
      \texttt{FP8 E5M2},\ \texttt{MXFP8 E4M3},\ \texttt{MXINT8}\},
  \end{equation*}
  
indexed by integer codes $0,1,2,3,4$ in the runtime \texttt{LUT}. The first three are conventional per-tensor formats; the last two are microscaling formats with block size 32 and a shared 8-bit exponent.

The weight palette is bit-width dependent:
\begin{align*} 
    \mathcal{F}_w^{(8)} \;&=\; \{\texttt{INT8-sym},\ \texttt{INT8-asym},\
    \texttt{MXINT8}\},\\
    \mathcal{F}_w^{(4)} \;&=\; \{\texttt{INT4-sym},\ \texttt{INT4-asym},\
    \texttt{NF4},\ \texttt{FP4 E2M1},\ \texttt{MXINT4},\ \texttt{MXFP4 E2M1}\}.
\end{align*}
$\mathcal{F}_w^{(4)}$ above is the full 4-bit palette; each architectural fork searches a subset of it (Section~\ref{subsec:archFork}), and Appendix~\ref{app:formats} describes each code.

We keep the palette small: the \rev{format} landscape has clear plateaus (Phases 1--3 of Appendix~\ref{app:phases}), and extra formats inside a plateau buy nothing while inflating the \texttt{LUT}. \texttt{MXINT8} is defined in the activation code space for symmetry with the weight palette but is never selected by the routing rules of Section~\ref{subsec:rthresh}; the remaining four are the smallest set covering all three phases plus the concatenation case. Alternatives to plain integer and floating-point coding exist in the broader compression literature \citep{lns_mac, positjj}, but we restrict the palette to formats that map onto upcoming hardware; Appendix~\ref{app:formats} describes each.

\subsection{Routing Thresholds}
\label{subsec:rthresh}

For each (layer, bucket) pair we apply the decision rule below, which \rev{replaces the argmin of \eqref{eq:format-argmin} with two robust statistics; selecting activation formats by calibration error directly is worse for generation quality (Appendix~\ref{app:rebuttal})}. \texttt{UNet} shortcut layers are pinned by topology before any statistic is consulted; the rest are routed by $\kappa$ and \texttt{SNR}:
\begin{align*}
  \text{If } l \text{ is a \texttt{UNet} shortcut layer:} &\quad
      f^{\star}_{l,b} = \texttt{MXFP8 E4M3}
      &&\text{(shortcut concatenation)}\\
  \text{Else if } \kappa_l(t) > 5 \text{ or } \mathrm{SNR}(b) < 0.2: &\quad
      f^{\star}_{l,b} = \texttt{FP8 E5M2}
      &&\text{(Phase 1, heavy tails)}\\
  \text{Else if } \kappa_l(t) \le 3 \text{ and } \mathrm{SNR}(b) > 2: &\quad
      f^{\star}_{l,b} = \texttt{INT8-asym}
      &&\text{(Phase 3, clean signal)}\\
  \text{Else:} &\quad
      f^{\star}_{l,b} = \texttt{FP8 E4M3}
      &&\text{(Phase 2, settling)}
\end{align*}

Here $\mathrm{SNR}(b) \equiv \mathrm{SNR}(t_b)$, evaluated at the bucket midpoint $t_b = b\,T/B + T/(2B)$, as in Algorithm~\ref{alg:calib}. The \texttt{SNR} gate overrides kurtosis by design: below $\tau_\mathrm{low}$ every non-shortcut layer is routed to \texttt{FP8 E5M2} (Table~\ref{tab:lutstats}), so per-layer routing acts in the mid- and high-\texttt{SNR} buckets. The constants $(\kappa_\mathrm{low}, \kappa_\mathrm{high}) = (3, 5)$ and $(\tau_\mathrm{low}, \tau_\mathrm{high}) = (0.2, 2.0)$ are fixed a priori rather than tuned per model: \rev{$\kappa_\mathrm{high} = 5$ is where \texttt{INT8} stops preserving the bulk of a heavy-tailed distribution -- under a generalized-Gaussian model with the shipped static scales, its bulk \texttt{SQNR} falls below \texttt{FP8 E4M3}'s between $\kappa = 4.5$ and $5.3$, whereas energy-weighted mean squared quantization error (\texttt{MSQE}) is dominated by a few outliers -- and $\kappa_\mathrm{low} = 3$ is a conservative bound below it (Appendix~\ref{app:rebuttal}),} and the \texttt{SNR} gates mark the macroscopic phases of the noise schedule. Empirically, the routing is insensitive to their exact values: shifting the kurtosis gates from $(3.0, 5.0)$ to $(3.5, 5.5)$ -- $+17\%$ and $+10\%$ respectively -- changes \texttt{FID} by $0.02$, below the $0.04$ run-to-run variation we observe between two independent calibrations of an identical configuration (Appendix~\ref{app:ablations}). \rev{A one-at-a-time grid over all four thresholds moves \texttt{FID} by at most $0.13$, within the spread across three calibration seeds (SD up to $0.14$), which is larger than this two-run estimate (Appendix~\ref{app:rebuttal}).}

\subsection{Calibration: Lockstep Sampling and LUT Construction}
\label{subsec:calib}

Calibration runs as a single pass on top of a \texttt{Q-Diffusion}-style uniform timestep sampler. We collect 128 prompts (\texttt{COCO} captions), bucket the schedule into $B = 10$ uniform-width windows, and generate one 20-step trajectory per prompt. During each forward pass, every Chameleon-instrumented layer accumulates running statistics into an $L \times B$ table, pooling the conditional and unconditional halves of each classifier-free-guidance batch; Algorithm~\ref{alg:calib} in the supplement gives the full pseudocode.

The accumulator stores raw power sums ($n$, $\sum x$, $\sum x^2$, $\sum x^3$, $\sum x^4$) plus a running min and max, from which $\kappa_l(t)$ follows in closed form. Memory is $O(1)$ in the number of prompts -- seven scalars per (layer, bucket) -- so \texttt{SDXL}'s 372 quantized layers at $B = 10$ need $26{,}040$ scalars, about $200$\,KB.
  
\subsection{Per-Channel Weight Selection}
\label{subsec:weightSel}

Weights are static, so their format selection runs offline and at finer granularity. For each output channel $c$ of each conv or linear, we round-trip $W_l[c, :, \ldots]$ through every format in the appropriate palette and pick the format with the highest quantization signal-to-noise ratio (\texttt{SQNR}) in dB:
\begin{equation}
  f^{\star}_{l,c} \;=\; \argmax_{f \in \mathcal{F}_w}\; 10 \log_{10}
  \frac{\mathbb{E}\!\left[W_l[c]^2\right]}
       {\mathbb{E}\!\left[\bigl(W_l[c] - \mathrm{DQ}_f(\mathrm{Q}_f(W_l[c]))\bigr)^2\right]}.
  \label{eq:weight-argmax}
\end{equation}

The per-channel grain matters: even within one \texttt{SDXL} conv layer, the kurtosis of weights along the output-channel axis varies enough that a tensor-wise choice would force a compromise no channel benefits from. It costs one byte of format metadata per output channel, about $1$\,MB across the \texttt{UNet}.
  
\subsection{The Architectural Fork}
\label{subsec:archFork}

A model loader inspects the network topology once and dispatches to one of three execution paths. The fork is the only branch in the framework; everything downstream is shared.

\paragraph{\texttt{UNet} diffusion path (\texttt{SDXL}):} Activations follow Section~\ref{subsec:rthresh} with static, ahead-of-time-calibrated scales (\texttt{Q-Diffusion} style). Layers matching the regex \texttt{up\_blocks.*.resnets.*.conv1} -- the conv layers immediately downstream of a shortcut concatenation -- are unconditionally routed to \texttt{MXFP8 E4M3}, bypassing the $\kappa$/\texttt{SNR} check, because their inputs are wide-tailed in a way that no per-tensor format can absorb. Weights follow Section~\ref{subsec:weightSel}.

\paragraph{Few-step path (\texttt{SDXL-Turbo}):} 
A one-step schedule has no timestep axis to route over, so this path inverts the division of labor: Chameleon owns the \emph{weights} and inherits the \emph{activations} wholesale. Activations use \texttt{MixDQ}'s calibrated static per-tensor scales together with its metric-decoupled mixed-precision assignment, unchanged. Weights are replaced by Chameleon's adaptive palette -- per-channel \texttt{SQNR} selection over $\mathcal{F}_w^{(8)}$ at 8 bits, and a per-layer $(g, f)$ search over $\{32, 64, 128, 192, 288\} \times \{\texttt{INT4-asym}, \texttt{NF4}, \texttt{FP4 E2M1}, \texttt{MXINT4}, \texttt{MXFP4 E2M1}\}$ at 4 bits (the \texttt{MX} formats carry one power-of-two scale per 32-element block, so $g$ does not change them and they are searched at $g = 32$) -- substituted for \texttt{MixDQ}'s symmetric fold at the point where the quantized weight tensor materializes (the nine up-block \texttt{conv\_shortcut} layers stay \texttt{FP16} at both bit-widths; Appendix~\ref{app:setup}). The weight palette alone therefore accounts for the improvement over \texttt{MixDQ} at both bit-widths, and \texttt{MixDQ}'s \texttt{BOS}-aware bypass becomes unnecessary under it (Appendix~\ref{app:ablations}), which also removes its batch-size-one inference constraint.

\paragraph{\texttt{DiT} path (\texttt{PixArt}-$\mathtt{\alpha}$):}
The \texttt{MX} shortcut logic is dropped -- \texttt{DiT}s do not have shortcut concatenations of the kind \texttt{UNet}s do. Two changes replace the shortcut rule:

(i) Weights use a joint $(g, f)$ search. For each linear layer we evaluate the \texttt{SQNR} -- input-aware at 4 bits, i.e.\ weighting each input channel's error by its activation energy $\mathbb{E}[x_i^2]$ -- of every $(g, f) \in \{32, 64, 128, 192, 288\} \times \{\texttt{INT4-asym}, \texttt{NF4}, \texttt{FP4 E2M1}\}$ tuple and pick the argmax. The search is restricted to the non-\texttt{MX} members of $\mathcal{F}_w^{(4)}$, since a searched $g$ already supplies the block granularity \texttt{MX} would contribute. Groups run along the \emph{input-channel} dimension, following \texttt{Q-DiT}'s observation that this is where \texttt{DiT} activations have their dominant variance.

(ii) Activations use \emph{macro-routing with dynamic per-sample scaling}: the format choice $f^{\star}_{l,b}$ from the \texttt{LUT} of \eqref{eq:format-argmin} is the macro decision, but the scale is computed dynamically from the current sample's range at inference rather than read from a calibrated table. This follows \texttt{Q-DiT}'s observation that \texttt{DiT} activations shift drastically across samples. Applied unchanged to \texttt{PixArt}-$\mathtt{\alpha}$, this rule produces a markedly different assignment from the \texttt{UNet}'s. At both bit-widths, \texttt{FP8 E5M2} wins in $97$--$100\%$ of the $282$ quantized linears in \emph{every} bucket and \texttt{INT8-asym} is never selected, so the three-phase progression of Appendix~\ref{app:phases} does not appear here. Since $\mathrm{SNR}(b)$ is large in the cleanest buckets, the gate that fires must be $\kappa_l(t) > 5$: \texttt{DiT} activations stay heavy-tailed along the entire schedule. The thresholds are identical across both architectures, with no per-model tuning, yet they track the architecture rather than reproducing the \texttt{UNet}'s answer.

\subsection{Implementation and Runtime}
\label{subsec:implRun}

The runtime path is a single \texttt{nn.Module} inserted before each quantizable layer; Algorithm~\ref{alg:runtime} gives the pseudocode and Appendix~\ref{app:algorithms} two reproducibility notes (\texttt{LUT} buffer registration and \texttt{FP8} range clamping). Here we say how our \texttt{MX} numbers should be read.

\texttt{MX} block formats are simulated by reshaping to blocks, computing the shared exponent, round-tripping each element, and reshaping back. The activation path follows the \texttt{OCP} specification exactly: the shared scale is the power-of-two $2^{\lfloor \log_2 \max_i |x_i| \rfloor}$ (an \texttt{E8M0} exponent), so the simulation matches conforming \texttt{MX} hardware. The \texttt{SDXL} weight-side implementation instead uses an unconstrained per-block scale, which is strictly more expressive than the specification allows (the few-step path's \texttt{MX} weights already use power-of-two scales, and the \texttt{DiT} path selects no \texttt{MX} weight format); the numbers we report for \texttt{MX} weight formats should be read accordingly. Appendix~\ref{app:mxconformance} measures what conformance would cost and how it changes the palette. \rev{Quality numbers use fake quantization throughout; real kernels for the \texttt{DiT} path, with exact Chameleon numerics on Ada and native \texttt{MX} on Blackwell, are reported in Appendix~\ref{app:rebuttal}.}

The whole framework is roughly $5.8$ kLOC of PyTorch. We will release the code and the per-model \texttt{LUT}s that reproduce Section~\ref{sec:experiments} exactly.

\section{Experiments}
\label{sec:experiments}

\subsection{Setup}
\label{subsec:expSetup}

We quantize \texttt{SDXL} base~1.0 \citep{podell2023sdxl}, \texttt{SDXL-Turbo} \citep{sauer2023adversarial}, and \texttt{PixArt}-$\mathtt{\alpha}$ \texttt{XL/2 1024-MS} \citep{chen2023pixart} through the \texttt{UNet}, few-step, and \texttt{DiT} paths of Section~\ref{subsec:archFork}, and re-run each family's prior \texttt{PTQ} schemes through one shared harness with identical prompts, seeds, and scoring. We report clean-\texttt{FID} \citep{parmar2022cleanfid} over 24{,}576 \texttt{COCO}-2014 validation captions \citep{lin2014coco} and \texttt{CLIP}-score \citep{radford2021clip} as a prompt-adherence check; resolutions and samplers differ per family (Table~\ref{tab:combined_results}), so \texttt{FID} is comparable within a backbone but not across. \rev{Quality results use fake quantization on a single A100; real-kernel latency and memory are reported in Appendix~\ref{app:rebuttal}.} Appendix~\ref{app:setup} gives the full setup.

\begin{table}[htbp]
      \centering
      \caption{Combined evaluation on \texttt{COCO}-2014 (24{,}576 prompts, one
      image per caption). Clean-\texttt{FID} is computed against the full val2014
      set and \texttt{CLIP}-score (\texttt{ViT-L/14}, cosine $\times 100$) against
      the matching captions; within each backbone and bit-width, the best quantized \texttt{FID} and the best quantized \texttt{CLIP} are in
      \textbf{bold} (\texttt{FP16} is the unbolded reference). \textsuperscript{*}Ours. In both \texttt{SDXL-Turbo} rows the nine up-block \texttt{conv\_shortcut} layers stay \texttt{FP16} (Section~\ref{subsec:archFork}).}
      \label{tab:combined_results}
      \renewcommand{\arraystretch}{1}

      \begin{tabular}{@{} l l l c c @{}}
          \toprule
          \textbf{Setup} & \textbf{Method} & \textbf{Bits} &
          \textbf{FID} $\downarrow$ & \textbf{CLIP} $\uparrow$ \\
          \midrule

          \multirow{7}{*}{\makecell[l]{\textbf{SDXL} \\ (50 steps, CFG 7.5,\\ $1024\times1024$)}}
          & \texttt{FP16} baseline               & W$_{16}$A$_{16}$  & 16.16          & 26.88 \\
          & \texttt{Q-Diffusion}                 & W$_{8}$A$_{8}$    & 14.70          & \textbf{26.77} \\
          & \texttt{PTQ4DM}                      & W$_{8}$A$_{8}$    & 14.65          & 26.68 \\
          & Chameleon\textsuperscript{*} & W$_{8}$A$_{8}$   & \textbf{14.23} & 26.70 \\
          \cmidrule(l){2-5}
          & \texttt{Q-Diffusion}                 & W$_{4}$A$_{8}$    & 21.41          & 26.27 \\
          & \texttt{PTQ4DM}                      & W$_{4}$A$_{8}$    & 21.82          & 26.21 \\
          & Chameleon\textsuperscript{*} & W$_{4}$A$_{8}$   & \textbf{14.43} & \textbf{26.64} \\
          \midrule

          \multirow{5}{*}{\makecell[l]{\textbf{SDXL-Turbo} \\ (1 step, no CFG,\\ $512\times512$)}}
          & \texttt{FP16} baseline                    & W$_{16}$A$_{16}$ & 21.63          & 26.63 \\
          & \texttt{MixDQ}                            & W$_{8}$A$_{8}$   & 21.43          & \textbf{26.64} \\
          & Chameleon\textsuperscript{*} & W$_{8}$A$_{8}$   & \textbf{20.93} & 26.62 \\
          \cmidrule(l){2-5}
          & \texttt{MixDQ}                            & W$_{4}$A$_{8}$   & 24.50          & 25.88 \\
          & Chameleon\textsuperscript{*} & W$_{4}$A$_{8}$   & \textbf{21.29} & \textbf{26.85} \\
          \midrule

          \multirow{5}{*}{\makecell[l]{\textbf{PixArt-$\alpha$} \\ (20 steps, CFG 4.5,\\ $1024\times1024$)}}
          & \texttt{FP16} baseline                    & W$_{16}$A$_{16}$ & 27.63          & 25.99 \\
          & \texttt{Q-DiT}                            & W$_{8}$A$_{8}$   & 27.69          & \textbf{25.99} \\
          & Chameleon\textsuperscript{*} & W$_{8}$A$_{8}$   & \textbf{24.33} & 25.81 \\
          \cmidrule(l){2-5}
          & \texttt{Q-DiT}                            & W$_{4}$A$_{8}$   & 23.53          & 25.73 \\
          & Chameleon\textsuperscript{*} & W$_{4}$A$_{8}$   & \textbf{22.24} & \textbf{26.09} \\

          \bottomrule
      \end{tabular}
  \end{table}

\subsection{Main Results}
\label{subsec:mainResults}


Table~\ref{tab:combined_results} reports the headline numbers for the three backbones.

In \texttt{SDXL}, Chameleon reaches $14.23$ \texttt{FID} at W$_{8}$A$_{8}$ and $14.43$ at W$_{4}$A$_{8}$, against \texttt{FP16} $16.16$ and the best prior results of $14.65$ (\texttt{PTQ4DM}) and $21.41$ (\texttt{Q-Diffusion}); the $7.0$-point margin at 4 bits is the largest in the table. \texttt{CLIP} slips from $26.88$ to $26.70$ and $26.64$, the latter the best of any quantized method by $0.37$. Both configurations share the same A$_{8}$ routing, so the $0.20$ \texttt{FID} difference is attributable to the weight palette alone.

In \texttt{SDXL-Turbo}, \texttt{FID} improves from $21.63$ in \texttt{FP16} to $20.93$ (W$_{8}$A$_{8}$) and $21.29$ (W$_{4}$A$_{8}$), with \texttt{CLIP} holding at $26.62$ and rising to $26.85$, the latter $0.22$ above the \texttt{FP16} reference. Because this fork inherits \texttt{MixDQ}'s activations unchanged, the margin over \texttt{MixDQ} -- $0.50$ \texttt{FID} at 8 bits, $3.21$ at 4 bits -- isolates the weight palette: the tighter the weight budget, the more \rev{the weight format and granularity choice} buys. \rev{At 4 bits the search selects group-wise \texttt{INT4-asym} for $99\%$ of layers, so on this backbone the gain comes from the format and granularity choice rather than from mixing many formats (Appendix~\ref{app:rebuttal}).}

In \texttt{PixArt}-$\mathtt{\alpha}$, Chameleon reduces \texttt{FID} from $27.63$ to $24.33$ at W$_{8}$A$_{8}$ and $22.24$ at W$_{4}$A$_{8}$, beating \texttt{Q-DiT} by $3.36$ and $1.29$ respectively, and exceeding its \texttt{CLIP} by $0.36$ at 4 bits. On the weight side the difference is the joint $(g, f)$ search: at 8 bits it settles on \texttt{INT8-asym} everywhere but at $g = 32$ rather than \texttt{Q-DiT}'s $128$ (\texttt{MXINT8} is never selected), and at 4 bits on $g = 32$ throughout, with \texttt{NF4} -- unavailable to an \texttt{INT}-only baseline -- for 69 of the 282 linears.

Which axes the palette uses differs by backbone, as the statistics dictate: \texttt{SDXL} uses all of them (per-channel weights, per-layer activations, three temporal regimes, the topological \texttt{MXFP8} pin); \texttt{PixArt}-$\mathtt{\alpha}$ uses the weights and a near-uniform activation assignment (\texttt{FP8 E5M2} in $97$--$100\%$ of linears); \texttt{SDXL-Turbo}, whose single step leaves no timestep axis, uses the weight palette alone. Appendix~\ref{app:ablations} separates the activation axes on \texttt{SDXL}: against the best single fixed format, the per-layer choice is worth $0.82$ \texttt{FID} and the timestep axis a further $0.12$\rev{, which is within calibration-seed variation at A$_{8}$; at 6-bit activations a five-bucket table gives the best result (Appendix~\ref{app:rebuttal})}.

Two patterns hold across all three backbones. Every Chameleon configuration scores \texttt{FID} below its \texttt{FP16} reference, and on \texttt{SDXL} and \texttt{PixArt}-$\mathtt{\alpha}$ the 4-bit model is competitive with or better than the 8-bit model. We do not have a validated account of either: a low-pass explanation holds on \texttt{PixArt}-$\mathtt{\alpha}$ but not on the \texttt{UNet}s, as we discuss in Appendix~\ref{app:regularisation} of the supplement. A qualitative comparison at W$_{4}$A$_{8}$ is provided in Figure~\ref{fig:36images} in Appendix~\ref{app:qualitative}.

\subsection{Calibration Cost}
\label{subsec:calibCost}

Chameleon's calibration involves no gradient updates, no auxiliary networks, and no integer program of our own -- the few-step path excepted, and only because it consumes \texttt{MixDQ}'s precomputed mixed-precision assignment unchanged. Calibration takes at most 45 minutes on a single A100 ($45$ minutes for \texttt{SDXL}, $7$ for \texttt{PixArt}-$\mathtt{\alpha}$ and $1$ for \texttt{SDXL-Turbo}) and under an hour for all three backbones combined (Table~\ref{tab:calib_cost} in Appendix~\ref{app:algorithms}). The routing hyperparameters -- the kurtosis and \texttt{SNR} thresholds and the bucket count $B$ -- are fixed across every model, with no per-model tuning; only the palette is specialized per architecture, and structurally so (Section~\ref{subsec:archFork}).

\section{Discussion \& Limitations}
\label{sec:discussion}

\rev{Our quality numbers come from fake quantization. Real-kernel measurements so far cover the \texttt{DiT} path (Appendix~\ref{app:rebuttal}), where end-to-end gains are bounded by unquantized attention; kernels for the \texttt{UNet} paths remain future work.} The evaluation is limited to image generation; extending the routing to video \texttt{DiT}s, whose temporal-attention sensitivities are uncharacterized, is a natural next step, and sub-4-bit (W$_{2}$/W$_{1}$) regimes would require quantization-aware training rather than \texttt{PTQ}. The palette also assumes its formats exist in hardware: \texttt{FP8} and \texttt{MX} need Hopper- or Blackwell-class support, and on \texttt{INT}-only \texttt{NPU}s it reduces to its \texttt{INT} members. Per-channel mixed formats likewise do not map to one native \texttt{GEMM} tile; they would run as same-format sub-\texttt{GEMM}s after a channel permutation, or with weights dequantized in registers as existing W$_{4}$A$_{8}$ kernels do. Appendix~\ref{app:discussion} expands these points, adding the per-channel format-metadata overhead at extreme model scale and calibration generalization to out-of-distribution prompts.

\section{Conclusion}
\label{sec:conclusion}

We argued that at a fixed bit-width the \emph{number format} is not a design constant but a discrete variable, and that it should be chosen per tensor: per weight channel, per layer, and -- where a timestep axis exists -- \rev{optionally} per timestep bucket. The argument is structural: a format is priced for a distribution, and the distributions inside a denoising network differ across channels, across layers, and along the schedule, sliding from heavy-tailed, noise-dominated tensors at $t \to T$ to tightly clustered, structured signals at $t \to 0$, so no single format in the standard 8-bit palette is correctly priced for all of them. Tuning scales cannot \rev{fully} fix a format mismatch.

Chameleon makes that choice adaptive. Two ahead-of-time statistics -- empirical kurtosis and the closed-form diffusion \texttt{SNR} -- populate a format lookup table that the runtime indexes with one tensor read, and a single architectural fork specializes the routing to \texttt{UNet}, few-step, and \texttt{DiT} backbones. Across all three backbones and both bit-widths, the resulting quantized configurations reduce \texttt{FID} below the \texttt{FP16} reference while holding \texttt{CLIP}-score within $0.24$ of it, and lead every quantized baseline on \texttt{CLIP} at W$_{4}$A$_{8}$ (Table~\ref{tab:combined_results}). The framework is composable rather than competitive: \texttt{Q-DiT}'s group-wise weights and dynamic per-sample scaling consume Chameleon's \texttt{LUT} unmodified, and \texttt{MixDQ}'s activation pipeline composes with our weight palette so cleanly that its \texttt{BOS}-aware bypass is no longer required.

The natural next steps -- \texttt{Q-Sched}-style sampler adaptation, \rev{extending the real-kernel results of Appendix~\ref{app:rebuttal} to attention and to Hopper}, video \texttt{DiT}s, and sub-4-bit \texttt{QAT} -- are independent and additive.

\newpage
\bibliography{references, swarm}
\bibliographystyle{iclr2027_conference}


\newpage
\appendix
\section*{Supplementary Materials}

\noindent\textbf{Table of Contents}
\medskip

\begingroup
\renewcommand{\arraystretch}{1.15}
\begin{tabular}{@{}p{0.04\linewidth}p{0.85\linewidth}r@{}}

\multicolumn{3}{@{}l}{\textbf{Appendix}} \\[0.5ex]

A.1 & Extended Related Work \dotfill                     & \pageref{app:related}   \\
A.2 & Diffusion Basics \dotfill                        & \pageref{app:diffusion} \\
A.3 & Activation Phases for SDXL \dotfill                       & \pageref{app:phases} \\
A.4 & Experimental Setup \dotfill                        & \pageref{app:setup}         \\
A.5 & Architectural-Fork Diagrams \dotfill               & \pageref{app:fork_diagrams}   \\
A.6 & CLIP-Score Computation \dotfill                    & \pageref{app:clip_scaling}    \\
A.7 & Algorithm Pseudocode \dotfill                      & \pageref{app:algorithms}      \\
A.8 & On Chameleon FID below the FP16 Reference \dotfill & \pageref{app:regularisation} \\
A.9 & Ablations \dotfill                                 & \pageref{app:ablations} \\
A.10 & Qualitative Samples \dotfill                       & \pageref{app:qualitative}     \\
A.11 & Number-Format Details \dotfill                     & \pageref{app:formats}         \\
A.12 & MX Conformance and Its Cost \dotfill              & \pageref{app:mxconformance}      \\
A.13 & Extended Discussion and Limitations \dotfill      & \pageref{app:discussion}      \\
A.14 & Additional Experiments (Rebuttal) \dotfill & \pageref{app:rebuttal} \\

\multicolumn{3}{c}{} \\[-1ex] 

\multicolumn{3}{@{}l}{\textbf{Statements}} \\[0.5ex]

B.1 & AI Use Statement                             \dotfill  & \pageref{stat:ai}      \\
B.2 & Ethics Statement                             \dotfill  & \pageref{stat:ethics}      \\
B.3 & Reproducibility Statement                    \dotfill  & \pageref{stat:repro}      \\
B.4 & Acknowledgments                              \dotfill  & \pageref{stat:ack}      \\
\end{tabular}
\endgroup

\newpage

\section{Appendix}

\subsection{Extended Related Work}
\label{app:related}

\paragraph{Diffusion-model PTQ:} \texttt{Q-Diffusion} \citep{li2023qdiffusion} was the first \texttt{PTQ} scheme to recognize that the activation distributions inside a denoising \texttt{UNet} are not stationary across the diffusion schedule. Their fix has two parts: a \textit{uniform-timestep sampler} during calibration that visits the entire schedule rather than a single timestep, and a \textit{split quantizer} for the \texttt{UNet}'s shortcut layers, which separately quantize the deep and shallow halves of the concatenated tensor before applying a fused matmul. \texttt{PTQ4DM} \citep{shang2023ptq4dm} sharpened the timestep handling by partitioning the schedule into buckets and accumulating per-bucket calibration statistics, but kept the format fixed at \texttt{INT8}. Both methods take the format itself as a given and tune scales around it. Two later lines of work stay within \texttt{INT} but attack the error differently: \texttt{PTQD} \citep{he2023ptqd} models the quantization error as correlated noise and corrects it during sampling, while \texttt{EfficientDM} \citep{he2024efficientdm} recovers low-bit quality by fine-tuning \texttt{LoRA} weights, which places it closer to \texttt{QAT} than to \texttt{PTQ}.

\paragraph{Few-step and consistency-model PTQ:} Few-step distilled models -- whether obtained by consistency distillation, as in Latent Consistency Models \citep{luo2023lcm}, or by adversarial distillation, as in \texttt{SDXL-Turbo} \citep{sauer2023adversarial} -- are particularly hostile to quantization, because each of their one to four denoising steps carries a much larger fraction of the total semantic burden than a single step in a 50-step \texttt{DDIM} trajectory; quantization noise that an iterative model would absorb statistically becomes visible in the output. \texttt{MixDQ} \citep{zhao2024mixdq} addressed this by combining \emph{metric-decoupled} mixed precision -- separately measuring image-quality and text-alignment sensitivity per layer and assigning bit-widths via integer programming -- with a surgical \emph{\texttt{BOS}-aware quantization} (\texttt{BAQ}) that exempts the begin-of-sentence position of the text embedding from activation quantization in the cross-attention $K$ and $V$ projections, substituting a pre
computed \texttt{FP16} output. \texttt{Q-Sched} \citep{frumkin2025qsched} takes a different route, leaving the quantized weights frozen and learning a single pair of scalar preconditioning coefficients, applied to the latent and to the quantized noise prediction at every step, that warps the integration trajectory to compensate for the quantization-induced drift in the vector field. Our few-step fork applies Chameleon's weight palette over \texttt{MixDQ}'s calibrated activation pipeline, leaving the activation side entirely to \texttt{MixDQ}: with a single denoising step the timestep axis degenerates, so the routing of Section~\ref{subsec:archFork} applies to weights only; notably, its \texttt{BOS}-aware bypass becomes unnecessary under Chameleon's weight palette -- re-enabling a faithful equivalent changes \texttt{FID} by less than $0.15$ (Appendix~\ref{app:ablations}) -- which also lifts \texttt{MixDQ}'s batch-size-one inference constraint. \texttt{Q-Sched} is concurrent with and complementary to Chameleon.
 
\paragraph{DiT quantization:} Diffusion transformers exhibit large variance across the \emph{input} channel dimension of their linear-layer activations rather than output channels, which means standard tensor or output-channel-wise quantization underflows the small-magnitude channels and crushes the model. \texttt{Q-DiT} \citep{chen2024qdit} introduced \emph{group-wise} weight quantization with an evolutionary search over group sizes $g \in \{32, 64, 128, 192, 288\}$, and \emph{dynamic} on-the-fly scale and zero-point computation for activations because pre-calibrated scales drift unacceptably across samples. \texttt{Q-DiT}, again, fixes \texttt{INT} as the only number system. Our \texttt{DiT} fork keeps \texttt{Q-DiT}'s group search and dynamic micro-scaling, but generalizes the search space from group-only to $(g, f)$ joint over group size and format choice from $\{\texttt{INT4},\, \texttt{NF4},\, \texttt{FP4}\}$ at 4 bits or $\{\texttt{INT8\;sym},\, \texttt{INT8\;asym},\, \texttt{MXINT8}\}$ at 8 bits. \texttt{PTQ4DiT} \citep{wu2024ptq4dit} instead balances channel salience to reach W$_{4}$A$_{8}$ on \texttt{DiT}s, and \texttt{FP4DiT} \citep{chen2025fp4dit} quantizes \texttt{PixArt}-$\mathtt{\alpha}$ with floating-point rather than integer formats at W$_{4}$A$_{6}$. Neither is re-run in our harness\rev{; \texttt{ViDiT-Q} \citep{zhao2025viditq} is run in its own pipeline with our prompts, seeds and scoring and compared relative to its own \texttt{FP16} reference (Appendix~\ref{app:rebuttal})}, so Table~\ref{tab:combined_results} compares against the per-architecture baseline each family established.
 
\paragraph{Microscaling formats:} The \texttt{OCP} \texttt{MX} specification \citep{ocp2023mx, rouhani2023mx} standardized a family of block-shared-exponent narrow formats -- \texttt{MXFP8}, \texttt{MXINT8}, \texttt{MXFP6}, \texttt{MXFP4}, \texttt{MXINT4} -- in which each block of 32 elements carries one shared 8-bit exponent and the elements themselves are stored in a narrow integer or floating-point body. Our fake-quantization implements it directly in PyTorch (Section~\ref{subsec:implRun}). The hardware payoff is that the shared exponent absorbs intra-tensor magnitude variation without any per-channel scale infrastructure. To our knowledge, Chameleon is the first \texttt{PTQ} scheme for diffusion models to use \texttt{MX} formats \emph{adaptively}, routing concatenation-bearing \texttt{UNet} shortcut layers to \texttt{MXFP8} by topology.

\paragraph{LLM quantization, in passing:} \texttt{GPTQ} \citep{frantar2023gptq}, \texttt{AWQ} \citep{lin2023awq} and \texttt{SmoothQuant} \citep{xiao2023smoothquant} established the modern playbook for \texttt{LLM} \texttt{PTQ}, and \texttt{SmoothQuant}'s idea of migrating activation outliers into the weight space inspired several diffusion follow-ups. Edge-deployment-oriented variants such as \texttt{EntroLLM} \citep{sanyal_entrollm} compose entropy-coded weight compression on top of post-training quantization to cut bandwidth on resource-constrained devices. Format selection at a fixed bit-width also originates in this line: \texttt{MoFQ} \citep{zhang2023mofq} chooses \texttt{INT} or \texttt{FP} per layer by quantization error for \texttt{LLM}s, and defers finer granularity to future work. Chameleon inherits that idea and extends it along three axes -- \rev{a palette that includes \texttt{NF4} and the \texttt{MX} formats, the architectural fork, and an optional timestep axis} -- none of which has a counterpart in \texttt{LLM} quantization, where the temporal structure Chameleon exploits is absent.

\subsection{Diffusion Basics}
\label{app:diffusion}

This appendix expands Section~\ref{subsec:diffBasics}. A continuous-time forward diffusion is the \texttt{SDE}
\begin{equation}
\mathrm{d}x_t = \mu(x_t, t)\,\mathrm{d}t + \sigma(t)\,\mathrm{d}w_t,
\end{equation}
which gradually corrupts a clean sample $x_0 \sim p_\mathrm{data}$ into Gaussian noise $x_T \sim \mathcal{N}(0, I)$. Sampling is performed by integrating the corresponding reverse-time probability-flow \texttt{ODE}
\begin{equation}
\mathrm{d}x_t = \big[\mu(x_t, t) - \tfrac{1}{2} \sigma(t)^2 \nabla_x \log p_t(x_t)\big]\,\mathrm{d}t,
\end{equation}
with the score given by a learned noise predictor, $\nabla_x \log p_t(x_t) \approx -\epsilon_\theta(x_t, t)/\sqrt{1-\bar\alpha_t}$. The variance-preserving discretization and the closed-form $\mathrm{SNR}(t)$ of \eqref{eq:snr} are given in the main text; $\mathrm{SNR}(t)$ falls monotonically across five to eight orders of magnitude along the schedule; for \texttt{SDXL}'s scaled-linear discretization it runs from $\sim\!10^{3}$ at $t = 0$ to $\sim\!10^{-2}$ at $t = T$, and for \texttt{PixArt}-$\mathtt{\alpha}$'s linear schedule from $\sim\!10^{4}$ to $\sim\!10^{-4}$.

For a diffusion transformer (\texttt{DiT}) the architecture changes -- convolutional \texttt{UNet} blocks become stacks of self-attention and feed-forward layers -- but the temporal structure of the activations is the same. Few-step distilled models collapse the schedule to one to four sampler steps; each step is still indexed by an \texttt{SNR} value, but at a single sampler step (\texttt{SDXL-Turbo}) the timestep axis degenerates and Chameleon's routing applies to weights only (Section~\ref{subsec:archFork}).

\subsection{Activation Phases for SDXL}
\label{app:phases}

This appendix gives an illustrative worked example of the activation statistics summarized in Section~\ref{subsec:assymWvsAct}. Concretely, we instrument a fully trained \texttt{SDXL UNet}, run 128 calibration prompts through a 20-step denoising schedule, and record the empirical sample kurtosis
\begin{equation}
\label{eq:kurt}
\kappa_l(t) = \frac{\mathbb{E}[(x_l - \mu_l)^4]}{(\mathbb{E}[(x_l - \mu_l)^2])^2}
\end{equation}
for the input activations of each of the $372$ quantized convolution and linear layers -- attention Q/K/V projections and the stem/head convolutions are kept in \texttt{FP16} and excluded -- at each of ten timestep buckets. Together with the diffusion signal-to-noise ratio $\mathrm{SNR}(t)$, three patterns emerge consistently across the instrumented layers and define the regimes Chameleon routes between. Table~\ref{tab:lutstats} (Appendix~\ref{app:lutstats}) reports the per-bucket format distribution behind the percentages quoted below, and Table~\ref{tab:ab_routing} ablates the two statistics that separate the phases.

\paragraph{\textbf{Phase 1} (\textbf{low} \texttt{SNR}, $t \to T$):} $\kappa_l(t) > 5$ for the majority of layers. Although the network input is dominated by the additive Gaussian noise $\sqrt{1-\bar\alpha_t}\,\epsilon$ -- whose kurtosis is exactly $3$ -- the internal activations are markedly heavy-tailed: normalization layers and attention rescale the noise-dominated features unevenly, producing sparse large outliers. \texttt{INT8} uniform quantization underperforms here because those outliers force the scale wide and the central bins go empty; the calibrated router assigns \texttt{FP8 E5M2} to $97\%$ of layers in the final bucket.

\paragraph{\textbf{Phase 2} (\textbf{mid} \texttt{SNR}):} As coarse structure forms, $\kappa_l(t)$ settles into the range $[3,5]$ for most layers. The distribution still has tails that \texttt{INT8} handles poorly, but the dynamic range has tightened by an order of magnitude. \texttt{FP8 E4M3}, with three mantissa bits and a $\pm 448$ range, is the closest match, and becomes the modal choice ($61\%$ of layers in the mid buckets).

\paragraph{\textbf{Phase 3} (\textbf{high} \texttt{SNR}, $t \to 0$):} Activations are now dominated by the structured signal; the distribution narrows toward Gaussian for the bulk of layers. \texttt{INT8}, with its $256$ uniform bins, is the highest-density 8-bit representation and \rev{preserves the bulk of the distribution better than the \texttt{FP8} formats (higher bulk \texttt{SQNR}, Appendix~\ref{app:rebuttal})} where both routing gates clear -- $\kappa_l \le 3$ and $\mathrm{SNR}(t) > 2$ -- which occurs for roughly one layer in ten in the cleanest bucket; \texttt{E4M3} remains the plurality format elsewhere.

The transitions are not synchronous across layers -- early-resolution convolution blocks reach Phase 3 sooner than mid-block projection layers -- which is why the routing decision is made per (layer, bucket) and not per bucket alone.

\subsection{Experimental Setup}
\label{app:setup}

Section~\ref{subsec:expSetup} of the main paper summarizes the setup; the full version follows.

\paragraph{Models:} We evaluated three publicly available pre-trained image generation models, none of which are fine-tuned: Stable Diffusion XL (\texttt{SDXL}) base~1.0 \citep{podell2023sdxl}, \texttt{SDXL-Turbo}, a one-step adversarially distilled variant of \texttt{SDXL} \citep{sauer2023adversarial}, and \texttt{PixArt}-$\mathtt{\alpha}$ \texttt{XL/2 1024-MS} \citep{chen2023pixart}. \texttt{SDXL} uses the \texttt{UNet} path of Section~\ref{subsec:archFork}; \texttt{SDXL-Turbo} uses the few-step path, which applies Chameleon's adaptive weight palette on top of \texttt{MixDQ}'s calibrated static activations; \texttt{PixArt}-$\mathtt{\alpha}$ uses the image \texttt{DiT} path.

\paragraph{Dataset:} All runs use the \texttt{COCO}-2014 validation set \citep{lin2014coco}: we generate 24{,}576 images, one per caption sampled from \texttt{captions\_val2014.json}, and compute clean-\texttt{FID} \citep{parmar2022cleanfid} against the full val2014 set and \texttt{CLIP}-score (\texttt{ViT-L/14}) against the matching captions \citep{radford2021clip}.\footnote{\texttt{CLIP}-score is reported as cosine similarity $\times\,100$, following convention, so the values in Table~\ref{tab:combined_results} correspond to cosine similarities of $0.257$--$0.269$ and are directly comparable to numbers reported elsewhere in the text-to-image literature. Scores are computed with \texttt{ViT-L/14} against the caption each image was actually generated from; Appendix~\ref{app:clip_scaling} documents the scoring path. We treat \texttt{FID} as the primary metric and \texttt{CLIP} as a prompt-adherence check, since \texttt{CLIP} measures text--image alignment rather than distributional fidelity.}

\paragraph{Reference Point:} For each backbone, we report the unquantized \texttt{FP16} model as reference and compare Chameleon at W$_{8}$A$_{8}$ and W$_{4}$A$_{8}$ against the per-architecture prior \texttt{PTQ} schemes: \texttt{Q-Diffusion} \citep{li2023qdiffusion} and \texttt{PTQ4DM} \citep{shang2023ptq4dm} for \texttt{SDXL}, \texttt{MixDQ} \citep{zhao2024mixdq} for \texttt{SDXL-Turbo}, and \texttt{Q-DiT} \citep{chen2024qdit} for \texttt{PixArt}-$\mathtt{\alpha}$. These methods are surveyed in Section~\ref{sec:related} and serve as the inspiration for the architectural fork of Section~\ref{subsec:archFork}; for the head-to-head comparison in Table~\ref{tab:combined_results} we additionally re-run each of them through our shared evaluation harness, with an identical prompt set, seeds, and scoring pipeline.

\paragraph{Few-step quantization coverage:} Both the \texttt{MixDQ} baseline and Chameleon's few-step path run through \texttt{MixDQ}'s released pipeline in fake-quantization mode. In that mode the pipeline quantizes only layers whose input is not split: the nine up-block \texttt{conv\_shortcut} layers, which \texttt{MixDQ} would otherwise split-quantize at the concatenation boundary (1280/640/320 decoder channels), keep \texttt{FP16} weights and activations at both W$_{8}$A$_{8}$ and W$_{4}$A$_{8}$, although \texttt{MixDQ}'s mixed-precision configuration assigns them bit-widths. Chameleon's weight palette is applied at the same point in the pipeline as \texttt{MixDQ}'s weight fold and so does not reach these layers either. The exception applies identically to both rows, so the few-step comparison in Table~\ref{tab:combined_results} is like-for-like, but each few-step configuration carries these nine additional \texttt{FP16} layers.

\paragraph{Hardware:} All runs use a single NVIDIA A100 80\,GB. Calibration and inference are both performed in fake-quant (simulated quantization in \texttt{FP16} arithmetic); we therefore report image quality \rev{in the main results; real-kernel latency and memory are reported separately in Appendix~\ref{app:rebuttal}}. Calibration cost is a property of the method rather than of the deployment, so we do report it (Table~\ref{tab:calib_cost}).

\paragraph{Hyperparameters:} $B = 10$ timestep buckets, 128 calibration prompts, no gradient updates anywhere. Inference uses published step counts and native resolutions: \texttt{SDXL} at 50 steps with \texttt{CFG} 7.5 at $1024\times1024$, \texttt{SDXL-Turbo} at 1 step with guidance disabled (\texttt{CFG} 0) at $512\times512$, and \texttt{PixArt}-$\mathtt{\alpha}$ at 20 steps with \texttt{DPM-Solver} \citep{lu2022dpmsolver} and \texttt{CFG} 4.5 at $1024\times1024$. Because resolutions and samplers differ per family, \texttt{FID} magnitudes are comparable within a backbone but not across backbones. Seeds are fixed at the global-index level so every method generates the same prompt set.

\begin{figure}[!t]
  \centering
  \includegraphics[width=\textwidth]{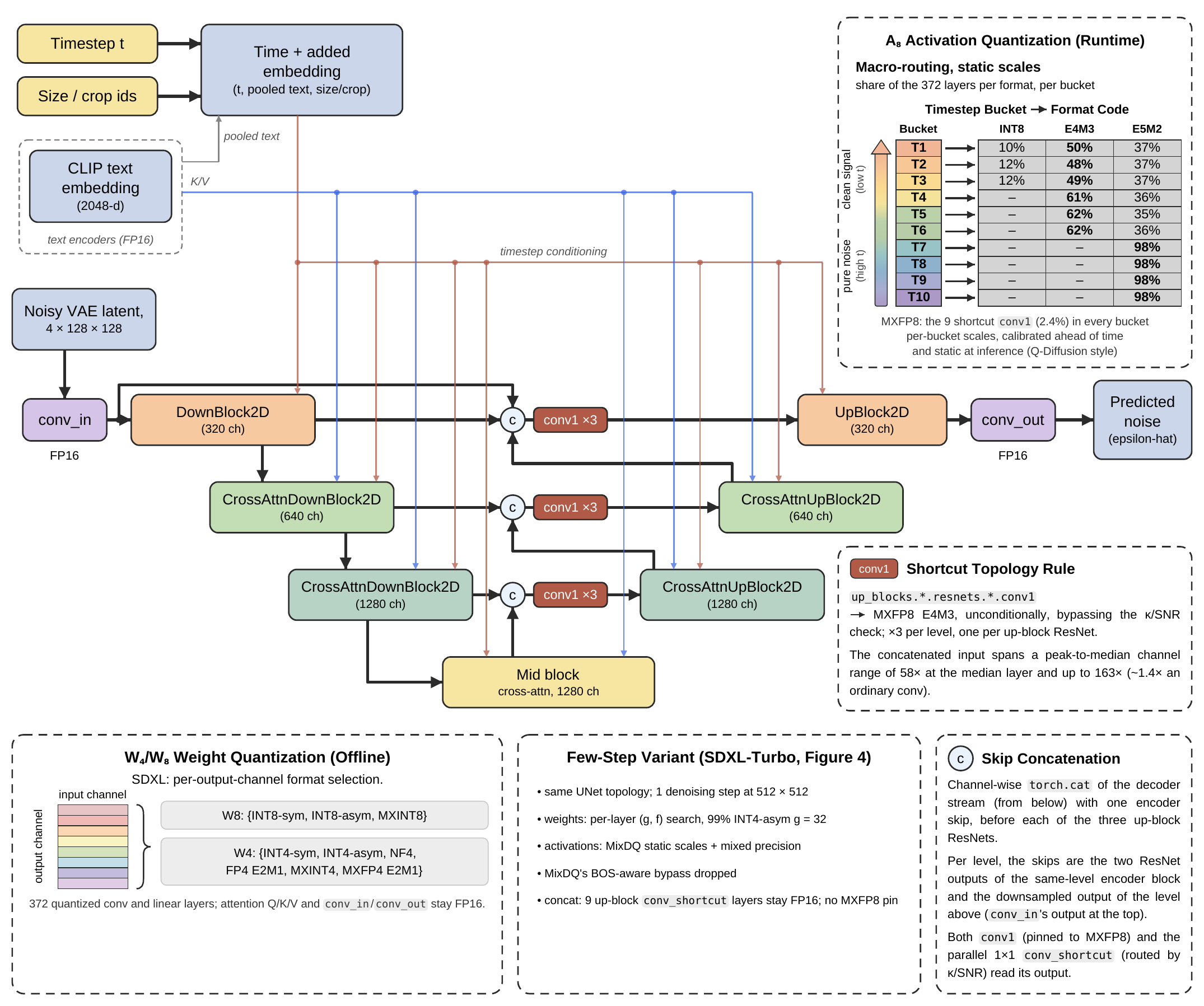}
  \caption{\textbf{The \texttt{UNet} execution path of Chameleon (\texttt{SDXL} base 1.0), with the few-step variant used for \texttt{SDXL-Turbo}.} The encoder (\texttt{DownBlock2D} and two \texttt{CrossAttnDownBlock2D}s) descends to a cross-attention mid block and the decoder climbs back up; at every level the concatenation $c$ joins the decoder stream with one encoder skip before each of the three up-block ResNets -- the encoder block's two ResNet outputs and the downsampled output of the level above (\texttt{conv\_in}'s output at the top). Activations follow macro-routing with static, ahead-of-time-calibrated per-bucket scales (\texttt{Q-Diffusion} style), in contrast to the \texttt{DiT} path's dynamic per-sample scales; the callout gives the measured share of the 372 quantized layers assigned each format in each bucket of the shipped \texttt{LUT} (Table~\ref{tab:lutstats}). The nine layers matching \texttt{up\_blocks.*.resnets.*.conv1} -- the $3\times3$ convolution that reads the concatenated tensor, three per level -- are routed to \texttt{MXFP8 E4M3} regardless of the kurtosis/\texttt{SNR} reading, because that tensor spans a peak-to-median per-channel magnitude range of $58\times$ at the median layer and up to $163\times$; the parallel $1\times1$ \texttt{conv\_shortcut} reads the same tensor and is routed normally. Weights use per-output-channel format selection (Section~\ref{subsec:weightSel}); attention Q/K/V projections, \texttt{conv\_in}/\texttt{conv\_out} and both text encoders stay in \texttt{FP16}. The few-step variant keeps the topology but runs one step at $512\times512$: Chameleon supplies the weight palette, activations use \texttt{MixDQ}'s static scales and mixed-precision assignment, \texttt{MixDQ}'s \texttt{BOS}-aware bypass is dropped (Table~\ref{tab:ab_fork}), and the nine up-block \texttt{conv\_shortcut} layers that read the concatenation stay in \texttt{FP16}, with no \texttt{MXFP8} pin (Section~\ref{subsec:archFork}; Appendix~\ref{app:setup}). \rev{Figure~\ref{fig:turbo} details the few-step path.}}
  \label{fig:unet}
\end{figure}

\begin{figure}[!t]
  \centering
  \includegraphics[width=\textwidth]{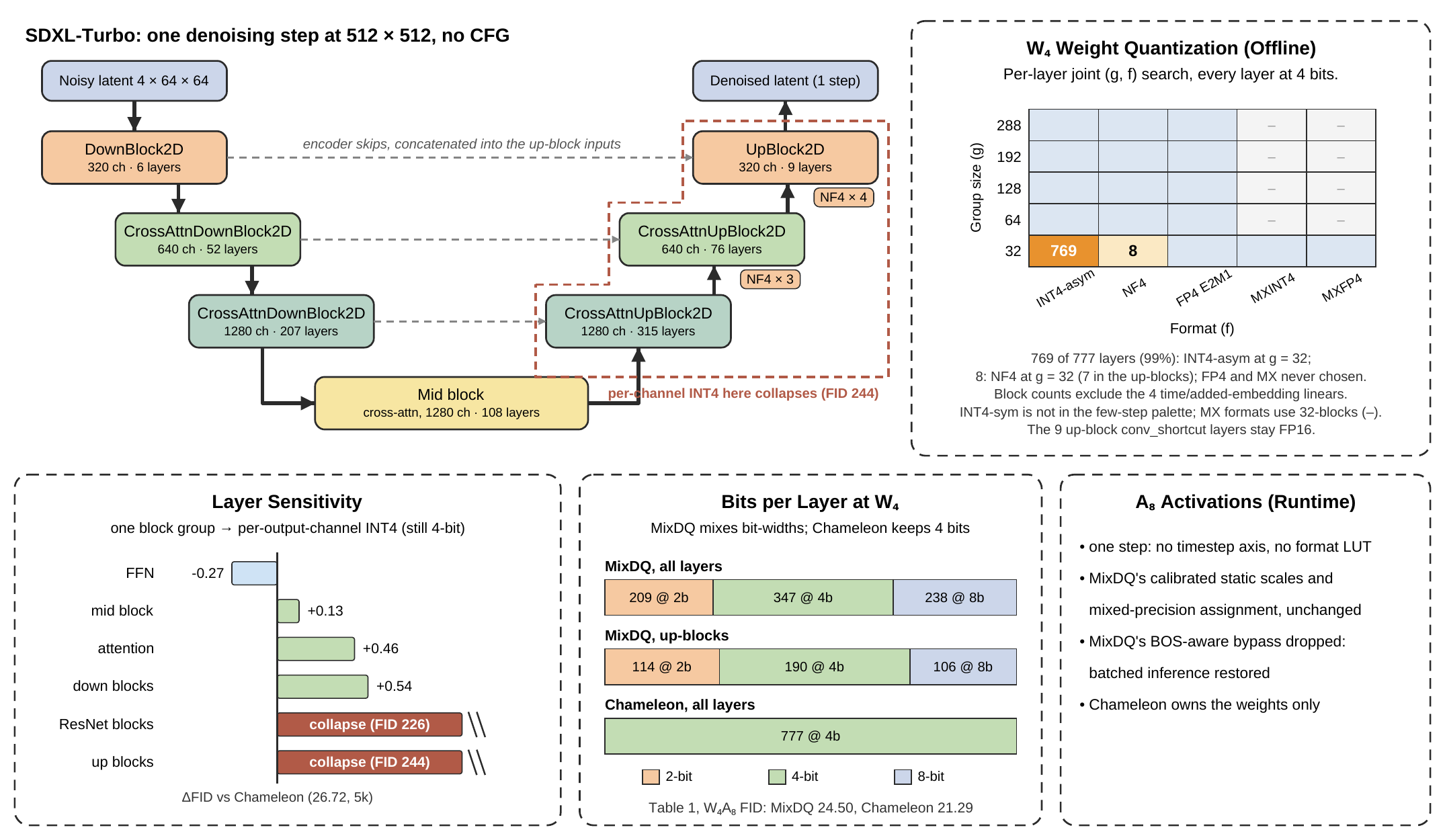}
  \caption{\rev{\textbf{The few-step path of Chameleon (\texttt{SDXL-Turbo}, one step at $512\times512$).} \emph{Weights:} a per-layer joint (group size, format) search keeps every one of the 777 quantized layers at 4 bits and selects \texttt{INT4-asym} at $g = 32$ for 769 of them and \texttt{NF4} at $g = 32$ for 8 (7 in the up-blocks). \emph{Sensitivity:} switching one block group to per-output-channel \texttt{INT4} costs at most $0.54$ \texttt{FID}, except in the up-block and ResNet layers that read skip-concatenated inputs, where generation collapses. \texttt{MixDQ} protects those layers with higher precision (106 of 410 up-block layers at 8 bits); Chameleon keeps them at 4 bits through its format and group-size choice. \emph{Activations:} \texttt{MixDQ}'s calibrated pipeline, unchanged except that the \texttt{BOS}-aware bypass is dropped. Numbers from Appendix~\ref{app:rebuttal} (5k images).}}
  \label{fig:turbo}
\end{figure}

\subsection{Architectural-Fork Diagrams}
\label{app:fork_diagrams}

\begin{figure}[!t]
  \centering
  
  \includegraphics[width=\textwidth]{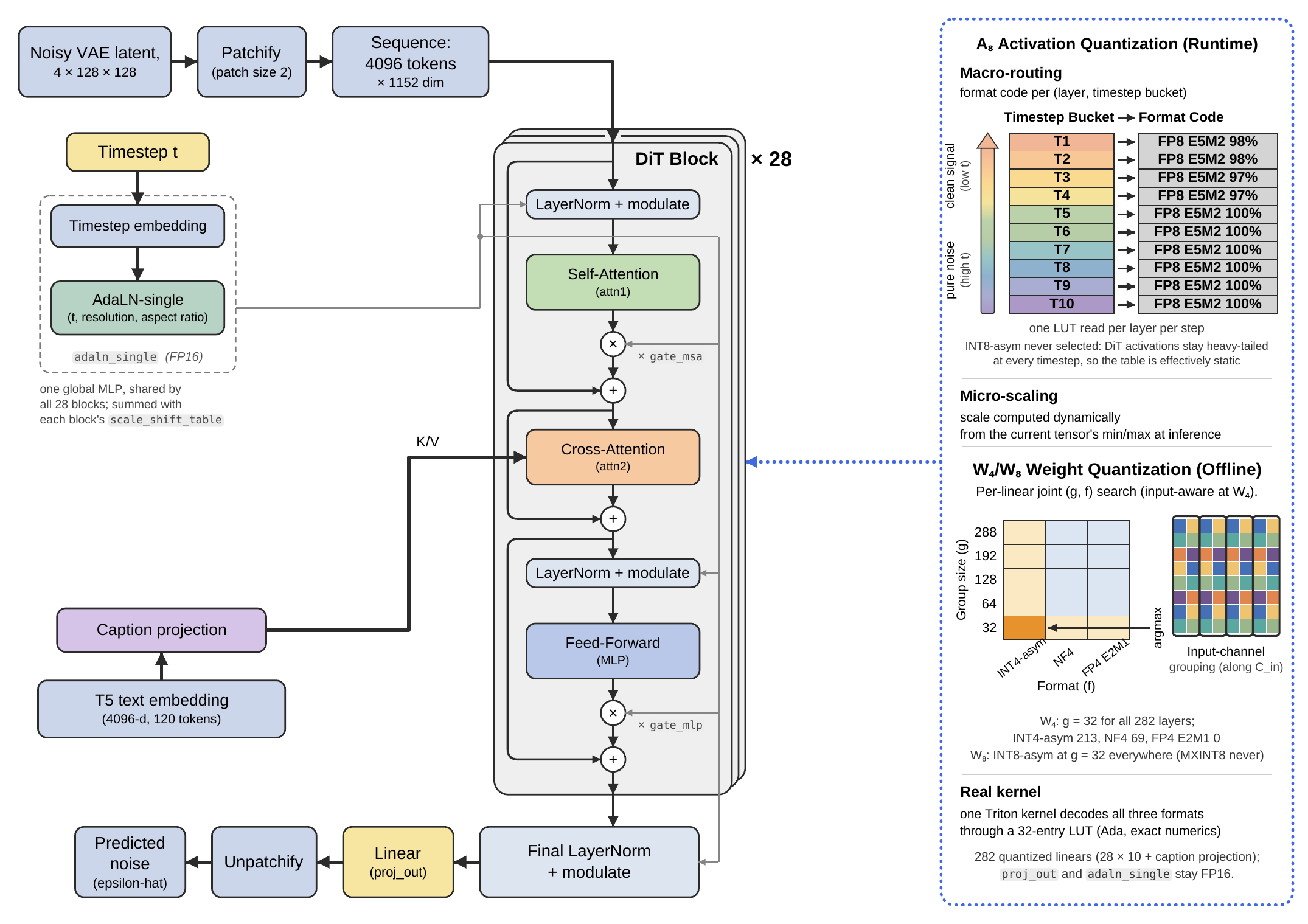}
  \caption{\textbf{The \texttt{DiT} execution path of Chameleon (\texttt{PixArt}-$\mathtt{\alpha}$ \texttt{XL/2}, 28 blocks).} There is no shortcut concatenation of the kind the \texttt{UNet} carries, so the topology-driven \texttt{MXFP8} rule of Section~\ref{subsec:archFork} is dropped. \emph{Weights:} every \texttt{nn.Linear} in the self-attention, cross-attention and feed-forward sub-layers, together with the two \texttt{caption\_projection} linears (282 in total; \texttt{proj\_out} and \texttt{adaln\_single} are kept in \texttt{FP16}) enters a joint $(g, f)$ search over group sizes $g \in \{32, 64, 128, 192, 288\}$ and formats $f \in \{\texttt{INT4-asym}, \texttt{NF4}, \texttt{FP4 E2M1}\}$ at 4 bits, scored by an input-aware \texttt{SNR} that weights each input channel's squared error by its activation energy from a 16-prompt pre-pass (it settles on $g = 32$ everywhere, with \texttt{INT4-asym} for 213 linears and \texttt{NF4} for 69); the search is per-linear, groups run along the \emph{input-channel} dimension, and it runs entirely offline. \emph{Activations:} macro-routing (the per-bucket format \texttt{LUT} of \eqref{eq:format-argmin}) composed with per-tensor dynamic per-sample scaling. The bucket-to-format assignment shown is the calibrated \texttt{LUT} for this model: \texttt{FP8 E5M2} wins in $97$--$100\%$ of the 282 linears in every bucket and \texttt{INT8-asym} is never selected, so the three-phase progression of Appendix~\ref{app:phases} does not occur on the \texttt{DiT}. Appendix~\ref{app:lutstats} reports the full statistics. The format is fixed at calibration time, but the scale is computed dynamically from the \emph{current} tensor's range at inference, following \texttt{Q-DiT}'s observation that \texttt{DiT} activations shift drastically across samples. \rev{The \texttt{W}$_8$ choice and the real-kernel note summarize Appendix~\ref{app:rebuttal}.}}
  \label{fig:dit-arch}
\end{figure}

This appendix collects the schematic diagrams for the execution paths of the architectural fork (Section~\ref{subsec:archFork} of the main paper). Figure~\ref{fig:dit-arch} shows the \texttt{DiT} path used for \texttt{PixArt}-$\mathtt{\alpha}$, and Figure~\ref{fig:unet} shows the \texttt{UNet} path used for \texttt{SDXL} together with the few-step modifications used for \texttt{SDXL-Turbo}\rev{, which Figure~\ref{fig:turbo} details}.


\subsection{CLIP-Score Computation}
\label{app:clip_scaling}

\paragraph{Caption alignment:} Each generated image must be scored against the caption it was generated from. Every per-backbone generator builds its prompt list as the flat list of raw annotations from \texttt{captions\_val2014.json} shuffled with a fixed seed ($42$), and writes image $i$ to \texttt{\{i:05d\}.png}; the scorer reconstructs the identical list and pairs \texttt{\{idx:05d\}.png} with \texttt{prompts[idx]}. Because the shuffle seed and the ordering convention are shared by all six generators and by every baseline, every configuration is scored against exactly the same captions in the same order, and the comparison across rows of Table~\ref{tab:combined_results} is like-for-like.

We flag this explicitly because an earlier version of our scorer built its caption list differently -- one caption per image identifier, unshuffled -- so that image $i$ was scored against an unrelated caption. The resulting values measured image-versus-random-caption similarity, clustered near cosine $0.075$, and did not preserve the ordering across configurations: the two most degraded ablation arms in Appendix~\ref{app:ablations} recorded the \emph{highest} scores. The released scorer reproduces the generators' caption list verbatim, and we verified equality across all $24{,}576$ indices before recomputing every number reported here.

\subsection{Algorithm Pseudocode}
\label{app:algorithms}

This appendix provides the full pseudocode for the algorithms summarized in the main paper. Algorithm~\ref{alg:calib} is the offline calibration loop referenced in Section~\ref{subsec:calib}; Algorithm~\ref{alg:runtime} is the runtime fake-quantization dispatch of the \texttt{UNet} and few-step paths referenced in Section~\ref{subsec:implRun}; Algorithm~\ref{alg:runtime_dit} is the corresponding dispatch for the \texttt{DiT} path, which replaces the calibrated scale with a per-sample one (Section~\ref{subsec:archFork}). 

Algorithms~\ref{alg:calib} and~\ref{alg:runtime_dit} both invoke a shared helper $\mathrm{ScaleZP}$, which derives the affine parameters of a format from an observed value range: \texttt{INT8-asym} takes $s = (\mathrm{hi} - \mathrm{lo})/255$ and $z = \mathrm{round}(-\mathrm{lo}/s)$; \texttt{FP8 E4M3} and \texttt{E5M2} take $s = \max(|\mathrm{lo}|, |\mathrm{hi}|)$ divided by $448$ and $57344$ respectively, with $z = 0$; the \texttt{MX} formats take $s = 1$, since the shared block exponent supplies the scaling. The difference between the two runtime paths is precisely \emph{when} this helper is called: the \texttt{UNet} path calls it once per (layer, bucket) at calibration time and stores the result, while the \texttt{DiT} path calls it on every forward pass from the current tensor's own range.

\begin{algorithm}[ht]
\caption{Chameleon \texttt{LUT} calibration}
\label{alg:calib}
\begin{algorithmic}[1]
\Require network $\mathbb{N}$ with $L$ instrumented layers, $B$ timestep
buckets, calibration prompt set $P = \{p_1, \ldots, p_K\}$, schedule
$t_T \rightarrow t_0$, cumulative products $\bar\alpha$
\Ensure format \texttt{LUT} $F \in \{0,\ldots,4\}^{L \times B}$, scale
\texttt{LUT} $S \in \mathbb{R}^{L \times B}$, zero-point \texttt{LUT}
$Z \in \mathbb{R}^{L \times B}$
\State initialize $n, \Sigma_1, \Sigma_2, \Sigma_3, \Sigma_4 \gets 0$ and
     $\mathrm{lo} \gets +\infty$, $\mathrm{hi} \gets -\infty$, all $L \times B$
\For{each prompt $p \in P$}
\State $x \gets$ random latent
\For{each step $t$ from $t_T$ to $t_0$}
  \State $b \gets \lfloor t / (T/B) \rfloor$
  \State $x \gets \mathbb{N}.\mathrm{step}(x, t, p)$
     \Comment{standard sampling pass}
  \For{each instrumented layer $l$}
    \State $X \gets$ cached input activation of $l$ at this step
    \State $n[l,b] \mathrel{+}= \mathrm{numel}(X)$; \;
           $\Sigma_k[l,b] \mathrel{+}= \sum_i X_i^{\,k}$ for $k = 1,\ldots,4$
    \State $\mathrm{lo}[l,b] \gets \min(\mathrm{lo}[l,b], \min_i X_i)$; \;
           $\mathrm{hi}[l,b] \gets \max(\mathrm{hi}[l,b], \max_i X_i)$
  \EndFor
\EndFor
\EndFor
\For{each $(l, b)$}
\State $\kappa[l,b] \gets m_4 / \sigma^4$, with $\sigma^2$ and $m_4$ recovered
       in closed form from $n, \Sigma_1, \ldots, \Sigma_4$
\State $t_b \gets b \cdot (T/B) + (T/B)/2$
       \Comment{bucket midpoint}
\State $F[l,b] \gets \mathrm{Route}\big(\kappa[l,b],\,
       \mathrm{SNR}(t_b),\, \mathrm{is\_shortcut}(l)\big)$
       \Comment{Section~\ref{subsec:rthresh}}
\State $S[l,b],\, Z[l,b] \gets \mathrm{ScaleZP}\big(F[l,b],\,
       \mathrm{lo}[l,b],\, \mathrm{hi}[l,b]\big)$
\EndFor
\end{algorithmic}
\end{algorithm}

\begin{algorithm}[ht]
\caption{\texttt{DynamicTimestepFakeQuantize.forward}$(x, t)$ --
\texttt{UNet} and few-step paths}
\label{alg:runtime}
\begin{algorithmic}[1]
\If{not \textit{enabled}} \Return $x$ \Comment{pre-calibration pass-through}
\EndIf
\State $b \gets \min\big(\lfloor t / \texttt{bucket\_size} \rfloor,\; B - 1\big)$
\State $(f, s, z) \gets \texttt{LUT}[\texttt{layer\_id}, b]$
     \Comment{one tensor-indexed read}
\If{$f = 0$} \Return $\mathrm{fq\_int8\_asym}(x, s, z)$
\ElsIf{$f = 1$} \Return $\mathrm{fq\_fp8\_e4m3}(x, s)$
\ElsIf{$f = 2$} \Return $\mathrm{fq\_fp8\_e5m2}(x, s)$
\ElsIf{$f = 3$} \Return $\mathrm{fq\_mxfp8\_e4m3}(x, \texttt{block\_size}{=}32)$
     \Comment{\texttt{MX}: block exponent, $s$ unused}
\ElsIf{$f = 4$} \Return $\mathrm{fq\_mxint8}(x, \texttt{block\_size}{=}32)$
     \Comment{in the code space; not emitted by Route}
\EndIf
\end{algorithmic}
\end{algorithm}

\begin{algorithm}[ht]
\caption{\texttt{ChameleonDiTDynamicQuantize.forward}$(x, t)$ --
\texttt{DiT} path}
\label{alg:runtime_dit}
\begin{algorithmic}[1]
\If{not \textit{enabled}} \Return $x$
\EndIf
\State $b \gets \min\big(\lfloor t / \texttt{bucket\_size} \rfloor,\; B - 1\big)$
\State $f \gets \texttt{format\_lut}[\texttt{layer\_id}, b]$
     \Comment{macro-routing: format only}
\State $\mathrm{lo}, \mathrm{hi} \gets \min(x), \max(x)$
     \Comment{micro-scaling: per-sample, per-forward}
\State $s, z \gets \mathrm{ScaleZP}(f, \mathrm{lo}, \mathrm{hi})$
\State \Return $\mathrm{fq}_f(x, s, z)$
\end{algorithmic}
\end{algorithm}


\paragraph{Implementation notes:} Two notes from Section~\ref{subsec:implRun} matter for reproducing Algorithms~\ref{alg:runtime} and~\ref{alg:runtime_dit}.

(i) The \texttt{LUT} and scale tables are registered as PyTorch \emph{buffers}, not parameters, so they move with the module across devices and serialize with the state dict. The format branch is data-dependent only in appearance: $f$ is constant for any given (layer, bucket), so a specializing compiler can in principle hoist the dispatch out of the inner loop. Our released implementation does not rely on this -- it reads the code from the buffer at each step -- and we leave the compiled runtime to future work. 

(ii) \texttt{FP8} round-trips clamp to the format's representable range \emph{before} the cast. Without clamping, values above $448$ map to \texttt{NaN} bit patterns in \texttt{E4M3} (which does not have $\pm\infty$), and values above $57344$ map to $\pm\infty$ in \texttt{E5M2}, which then becomes \texttt{NaN} through the $\infty - \infty$ that subsequent norm layers compute, and the model produces black images. This was the single most painful bug in our implementation, and we mention it because every quantization codebase we have read gets it wrong.

\paragraph{Calibration cost:} Table~\ref{tab:calib_cost} reports the end-to-end cost of Algorithm~\ref{alg:calib} for each backbone (Section~\ref{subsec:calibCost}).

\begin{table}[htbp]
  \centering
  \caption{End-to-end calibration cost on a single A100 80\,GB in fake-quant mode, with 128 calibration prompts and $B = 10$ buckets. The reported \texttt{LUT} size includes the per-(layer, bucket) format and scale tables; for \texttt{SDXL} and \texttt{PixArt}-$\mathtt{\alpha}$ the larger figure additionally includes the per-output-channel weight-format metadata. \texttt{SDXL-Turbo} is an order of magnitude cheaper because its one-step schedule leaves no timestep axis to route over: calibration reduces to a weight-only format search with no calibration forward passes (Section~\ref{subsec:archFork}). The three runs total 53\,m\,42\,s.}
  \label{tab:calib_cost}
  \renewcommand{\arraystretch}{1}
  \begin{tabular}{@{} c c c c @{}}
      \toprule
      \textbf{Model} & \textbf{Calib time} & \textbf{Peak GPU mem} &
      \textbf{\texttt{LUT} size} \\
      \midrule
      \texttt{SDXL} \texttt{UNet} & 45\,m\,16\,s & 11.44\,GB &
            \makecell{252\,KB (act.\ only) /\\ 18\,MB JSON of weight formats\\ (${\approx}1$\,MB at one byte per channel)} \\
      \texttt{SDXL-Turbo}         & 1\,m\,01\,s  &  9.63\,GB &
          157\,KB (single config file) \\
      \texttt{PixArt}-$\mathtt{\alpha}$             & 7\,m\,25\,s  & 18.27\,GB &
          93\,KB (per-layer $(g, f)$ + \texttt{LUT}) \\
      \bottomrule
  \end{tabular}
\end{table}

\subsection{On Chameleon FID below the FP16 Reference}
\label{app:regularisation}

Every Chameleon configuration in Table~\ref{tab:combined_results} reports a
\texttt{FID} below its unquantized \texttt{FP16} reference. The effect spans an
order of magnitude in size, from $-0.34$ on \texttt{SDXL-Turbo} at
W$_{4}$A$_{8}$ ($21.29$ against $21.63$) to $-5.39$ on \texttt{PixArt}-$\mathtt{\alpha}$ at
W$_{4}$A$_{8}$ ($22.24$ against $27.63$), and it appears on all three backbones
at both bit-widths. The reading is unusual -- one would normally expect the
unquantized model to dominate -- so we record what we can and cannot establish
about it.

\paragraph{The effect exceeds measurement noise:} \texttt{FID} is computed on
the full $24{,}576$-image generation against the $40{,}504$-image
\texttt{COCO}-2014 reference set. Two independently calibrated runs of an
identical configuration differ by $0.04$ \texttt{FID}
(Appendix~\ref{app:ablations}), so inversions spanning $0.34$ to $5.39$ are not
attributable to calibration or sampling variance\rev{; the larger spread across three calibration seeds (SD up to $0.14$, Appendix~\ref{app:rebuttal}) is also well below them}. Nor are they confined to a
single fork: they appear under three different weight-selection procedures
(per-output-channel \texttt{SNR} on \texttt{SDXL}, the adaptive palette over
\texttt{MixDQ} activations on \texttt{SDXL-Turbo}, and the joint $(g, f)$ search
on \texttt{PixArt}-$\mathtt{\alpha}$) and under both static and dynamic activation scaling.

\paragraph{The mechanism is not uniform smoothing:} A natural explanation is
that aggressive rounding attenuates the very-high-frequency texture that the
\texttt{InceptionV3} features underlying \texttt{FID} over-penalize relative to
perceptual quality -- an effect documented in the \texttt{GAN} compression
literature, where pruning and quantization occasionally improve \texttt{FID} for
this reason. We tested it directly by measuring the variance of the Laplacian
over $400$ prompt-matched images per configuration
(Table~\ref{tab:sharpness}). The explanation holds on \texttt{PixArt}-$\mathtt{\alpha}$, where
quantization reduces high-frequency energy from $1287$ to $1057$ -- and that
is also the backbone with the largest inversion. It does not hold for the two
\texttt{UNet} backbones, where the quantized models carry substantially
\emph{more} high-frequency content than their \texttt{FP16} references
($522 \rightarrow 662$ on \texttt{SDXL}, $439 \rightarrow 725$ on
\texttt{SDXL-Turbo}). Whatever produces the inversion there, it is not a
low-pass effect, and we do not have a validated account of it.

\paragraph{Bottom line for the main-text claims:} We report the inversions as
raw measurements and make no stronger claim. Because they are measured on a
single feature extractor with known frequency biases, and because we can rule
out the most obvious mechanism on two of three backbones, a reader skeptical of
\texttt{FID} as a single-number metric should treat the quantized
configurations as occupying the same neighborhood as their \texttt{FP16}
references rather than as strictly superior. We note that the corrected
\texttt{CLIP} scores of Appendix~\ref{app:clip_scaling} move in the same
direction on two of three backbones -- Chameleon exceeds the \texttt{FP16}
reference at W$_{4}$A$_{8}$ on \texttt{SDXL-Turbo} ($26.85$ against $26.63$) and
\texttt{PixArt}-$\mathtt{\alpha}$ ($26.09$ against $25.99$) -- but establishing a genuine quality
improvement over \texttt{FP16} would require a perceptual metric such as human
preference, which we have not run.


  \subsection{Ablations}
  \label{app:ablations}

  Each arm in this appendix generates 5{,}000 images from the same
  seed-42 \texttt{COCO} prompt subset and is scored with the clean-\texttt{FID}
  and \texttt{CLIP} pipeline of Appendix~\ref{app:setup}. The ablation study
  is self-contained: all arms share a common baseline generated from the same
  calibration cache, and that baseline ($19.39$) is not numerically identical to
  the \texttt{SDXL} W$_{4}$A$_{8}$ row of Table~\ref{tab:combined_results}, which
  is a sub-score of the 24{,}576-image run. Comparisons should therefore be read
  within this appendix rather than against the main table.

  Two independently calibrated runs of the \emph{same} configuration
  (\texttt{both\_default} in Table~\ref{tab:ab_routing} and $B=10$ in
  Table~\ref{tab:ab_buckets}) differ by $0.04$ \texttt{FID}. We take this as the
  empirical noise floor induced by calibration sampling, and treat differences at
  or below it as indistinguishable. \rev{Three calibration seeds per configuration (Appendix~\ref{app:rebuttal}) give a larger spread, an SD of up to $0.14$, so this two-run estimate understates calibration noise; differences below about $0.15$ should be read as unresolved.}

  Across the twenty arms, \texttt{FID} and \texttt{CLIP} are strongly
  anti-correlated ($r = -0.88$), as they should be when lower \texttt{FID} and
  higher \texttt{CLIP} both indicate better output. The two configurations that
  collapse -- \texttt{single\_int8} at $180.51$ \texttt{FID} and
  \texttt{macro\_only} at $174.42$ -- record the lowest \texttt{CLIP} scores in
  the study ($22.87$ and $15.91$ against $25.6$--$26.7$ for every healthy arm).
  We note this because it is a useful check on the scoring pipeline itself: under
  the caption misalignment described in Appendix~\ref{app:clip_scaling}, these same
  twenty arms gave $r = +0.93$, with the two collapsed configurations scoring
  \emph{highest}.

  \subsubsection{Routing Signals}

  \begin{table}[ht]
  \centering
  \caption{Contribution of each routing statistic (\texttt{SDXL}
  W$_{4}$A$_{8}$, 5{,}000 images).}
  \label{tab:ab_routing}
  \begin{tabular}{@{}llccc@{}}
  \toprule
  \textbf{Arm} & \textbf{Change} & \textbf{FID} $\downarrow$ & $\Delta$ &
  \textbf{CLIP} $\uparrow$ \\
  \midrule
  \texttt{both\_default}   & $\kappa$ and \texttt{SNR} (shipped)      & \textbf{19.39} & --     & \textbf{26.58} \\
  \texttt{both\_perturbed} & $\kappa$ gates $(3,5) \to (3.5,5.5)$     & 19.41 & $+0.02$ & 26.55 \\
  \texttt{kurt\_only}      & \texttt{SNR} gate disabled               & 19.60 & $+0.21$ & 26.59 \\
  \texttt{snr\_only}       & $\kappa$ frozen at 4 (mid-band)          & 19.68 & $+0.29$ & 26.38 \\
  \texttt{single\_e4m3}    & no routing, \texttt{FP8 E4M3} everywhere & 20.29 & $+0.90$ & 26.48 \\
  \bottomrule
  \end{tabular}
  \end{table}

  Routing is worth $0.90$ \texttt{FID} over the best single fixed format, and
  neither statistic alone recovers it: removing the \texttt{SNR} gate costs
  $0.21$ and freezing $\kappa$ costs $0.29$, so the two signals carry
  complementary information. The perturbation arm moves the kurtosis thresholds
  by $+17\%$ and $+10\%$ respectively and changes \texttt{FID} by $0.02$, just
  below the noise floor -- the thresholds are not tuned quantities.

  \subsubsection{Format Palette}

  \begin{table}[ht]
  \centering
  \caption{Contribution of individual formats (\texttt{SDXL} W$_{4}$A$_{8}$,
  5{,}000 images). \texttt{drop\_mxfp8} removes the only format the shortcut rule
  of Section~\ref{subsec:archFork} can select, and therefore disables that rule.}
  \label{tab:ab_palette}
  \begin{tabular}{@{}llccc@{}}
  \toprule
  \textbf{Arm} & \textbf{Change} & \textbf{FID} $\downarrow$ & $\Delta$ &
  \textbf{CLIP} $\uparrow$ \\
  \midrule
  \texttt{full\_palette}     & shipped                                       & \textbf{19.39} & --       & \textbf{26.58} \\
  \texttt{drop\_nf4}         & \texttt{NF4} removed (weights)                & 19.46  & $+0.07$   & 26.56 \\
  \texttt{drop\_mxfp8}       & \texttt{MXFP8} removed (shortcut rule off)    & 19.54  & $+0.15$   & 26.56 \\
  \texttt{drop\_mxfp4}       & \texttt{MXFP4 E2M1} removed (weights)         & 19.77  & $+0.38$   & 26.60 \\
  \texttt{single\_e4m3} & forced \texttt{E4M3}, shortcuts included      & 20.29  & $+0.90$   & 26.48 \\
  \texttt{single\_int8}      & forced \texttt{INT8-asym}, shortcuts included & 180.51 & $+161.12$ & 22.87 \\
  \bottomrule
  \end{tabular}
  \end{table}

  \rev{\texttt{MXFP4 E2M1} is clearly needed: dropping it costs $0.38$. The contributions of \texttt{NF4} ($0.07$) and of the shortcut rule's topology-driven \texttt{MXFP8} assignment on nine layers ($0.15$) are within calibration-seed noise (SD up to $0.14$, Appendix~\ref{app:rebuttal}), so these two arms do not resolve them.} The decisive row is \texttt{single\_int8}: forcing
  uniform integer quantization on every layer at every timestep raises
  \texttt{FID} by an order of magnitude ($19.39 \to 180.51$) and produces over-saturated,
  structurally intact but visually destroyed images. \rev{In a closely related configuration that keeps the nine shortcut layers in \texttt{MXFP8}, the collapse traces to inference-time activations exceeding the calibrated per-bucket \texttt{INT8} ranges: widening every range by $1.5\times$ gives \texttt{FID} $23.92$ under the 5k protocol of Appendix~\ref{app:rebuttal}, still $4.6$ \texttt{FID} behind format routing on the same platform (Table~\ref{tab:reb_mofq}). Scale tuning thus avoids the collapse but does not close the gap to choosing the format.}

  \subsubsection{Timestep Bucket Count}

  \begin{table}[ht]
  \centering
  \caption{Sensitivity to the number of timestep buckets (\texttt{SDXL}
  W$_{4}$A$_{8}$, 5{,}000 images). Each $B$ is calibrated independently.}
  \label{tab:ab_buckets}
  \begin{tabular}{@{}ccc@{}}
  \toprule
  $B$ & \textbf{FID} $\downarrow$ & \textbf{CLIP} $\uparrow$ \\
  \midrule
  1            & 19.47 & 26.58 \\
  5            & 19.41 & 26.57 \\
  10 (shipped) & \textbf{19.35} & \textbf{26.72} \\
  20           & 19.59 & 26.53 \\
  \bottomrule
  \end{tabular}
  \end{table}

  \texttt{FID} varies little with $B$: the total spread is $0.24$ over a
  twentyfold range of bucket counts. The shipped $B=10$ is the best setting on
  both metrics, but the margins are small. Collapsing the schedule to a single
  bucket ($B=1$, which reduces the router to one static format per layer) costs
  $0.12$ \texttt{FID} and $0.14$ \texttt{CLIP}; $B=5$ lies between the two; and
  $B=20$ is the worst of the four. The calibrated \texttt{LUT} explains the
  flatness (Appendix~\ref{app:lutstats}): on \texttt{SDXL} there are only three
  distinct temporal regimes, so once $B$ resolves them the extra buckets largely
  replicate their neighbors, while $B=20$ spreads the same 128 calibration
  trajectories over twice as many windows and the per-bucket moments get noisier.
  The two axes are therefore separable, and unequal: measured against the best
  single fixed format (\texttt{single\_e4m3}, $20.29$ in
  Table~\ref{tab:ab_routing}), the per-layer assignment at $B=1$ already recovers
  $0.82$ \texttt{FID} and the timestep axis adds the remaining $0.12$. Both
  tables' shipped runs differ by $0.04$ (below), so that split is approximate;
  what it does show is that per-tensor format choice carries the gain and
  temporal granularity refines it. \rev{Repeating $B \in \{1, 10\}$ with three calibration seeds each (Table~\ref{tab:reb_seeds}) reverses this single-seed ordering: $B = 1$ gives $19.24 \pm 0.05$ and $B = 10$ gives $19.48 \pm 0.14$. The $0.12$ timestep gain is therefore within calibration-seed variation at A$_8$; the per-layer choice carries the gain, and the timestep axis is an optional refinement that helps at 6-bit activations.}

  \subsubsection{Architectural-Fork Additions}

  \begin{table}[ht]
    \centering
    \caption{The per-family additions of Section~\ref{subsec:archFork}
    (5{,}000 images each). \texttt{macro\_only} freezes the per-sample activation scale after the first observation (format \texttt{LUT} still active); \texttt{micro\_only} keeps the dynamic scale but flattens the \texttt{LUT} to one format.}
    \label{tab:ab_fork}
    \begin{tabular}{@{}lllccc@{}}
    \toprule
    \textbf{Family} & \textbf{Arm} & \textbf{Change} & \textbf{FID} $\downarrow$ &
    $\Delta$ & \textbf{CLIP} $\uparrow$ \\
    \midrule
    \multirow{2}{*}{\makecell[l]{\texttt{SDXL-Turbo}\\W$_{8}$A$_{8}$}}
     & \texttt{baq\_off} & shipped                          & \textbf{26.78} & --      & 26.56 \\
     & \texttt{baq\_on}  & faithful \texttt{BAQ} re-enabled & 26.89          & $+0.11$ & \textbf{26.61} \\
    \midrule
    \multirow{4}{*}{\makecell[l]{\texttt{PixArt}-$\mathtt{\alpha}$\\W$_{4}$A$_{8}$}}
     & \texttt{macro+micro} & shipped                                 & \textbf{29.15} & --        & 25.65 \\
     & \texttt{micro\_only} & \texttt{LUT} flattened to \texttt{E4M3} & 29.84          & $+0.69$   & \textbf{25.83} \\
     & \texttt{macro\_only} & \makecell[tl]{scale frozen after\\first observation}
                                                                      & 174.42         & $+145.27$ & 15.91 \\
    \bottomrule
    \end{tabular}
  \end{table}

  The two \texttt{DiT} rows separate the fork's two mechanisms and show they are
  strongly asymmetric. Freezing the per-sample scale collapses generation
  entirely ($+145$ \texttt{FID}), confirming \texttt{Q-DiT}'s observation that
  \texttt{DiT} activations shift too much across samples for any static scale;
  dynamic per-sample scaling is load-bearing. Replacing the temporal \texttt{LUT} with
  a single fixed format costs $0.69$, a real but secondary contribution. That
  ordering -- per-sample adaptivity dominant, temporal granularity secondary --
  matches the bucket sweep on \texttt{SDXL}, where the timestep axis is worth
  $0.12$ \texttt{FID} against the per-layer choice's $0.82$ \rev{(the former within calibration-seed variation; see above)}.

  \texttt{CLIP} does not resolve these rows -- \texttt{micro\_only} scores $0.18$
  higher despite the worse \texttt{FID} -- which is consistent with the temporal
  \texttt{LUT} on \texttt{PixArt}-$\mathtt{\alpha}$ being close to uniform in the first place
  (Appendix~\ref{app:lutstats}).

  For the few-step fork, re-enabling a faithful equivalent of \texttt{MixDQ}'s
  \texttt{BOS}-aware bypass leaves quality essentially unchanged, and the two
  metrics disagree on its sign: \texttt{FID} rises by $0.11$ while \texttt{CLIP}
  rises by $0.05$. Neither movement supports a benefit from the bypass. What it
  does carry is a cost -- \texttt{MixDQ}'s \texttt{BAQ} substitutes a precomputed
  output for position $0$ of the batch, which forces batch-size-one inference.
  Under Chameleon's adaptive weight palette the \texttt{BOS} outlier no longer
  requires a bespoke mechanism to absorb it, so we drop the bypass and recover
  batched inference at no measurable quality cost.

  \subsubsection{What the Calibrated LUT Actually Selects}
  \label{app:lutstats}

  \begin{table}[ht]
  \centering
  \caption{Per-bucket activation-format distribution in the shipped
  \texttt{LUT}s ($B = 10$). \texttt{SDXL}: 372 quantized layers;
  \texttt{PixArt}-$\mathtt{\alpha}$: 282. Bucket mean $\mathrm{SNR}(t)$ is computed from each
  model's own noise schedule. Bucket $b$ is labelled T$_{b+1}$ in Figures~\ref{fig:unet} and~\ref{fig:dit-arch}.}
  \label{tab:lutstats}
  \small
  \begin{tabular}{@{}crcccc|rcc@{}}
  \toprule
  & \multicolumn{5}{c|}{\textbf{\texttt{SDXL}} (scaled-linear)} & \multicolumn{3}{c}{\textbf{\texttt{PixArt}-$\mathtt{\alpha}$} (linear)} \\
  $b$ & \texttt{SNR} & \texttt{INT8} & \texttt{E4M3} & \texttt{E5M2} & \texttt{MXFP8} & \texttt{SNR} & \texttt{E4M3} & \texttt{E5M2} \\
  \midrule
  0 & 57.7  &  9.9\% & 50.3\% & 37.4\% & 2.4\% & 304.5 & 2.5\% & 97.5\% \\
  1 & 5.14  & 12.1\% & 48.4\% & 37.1\% & 2.4\% & 4.18  & 2.5\% & 97.5\% \\
  2 & 2.13  & 11.6\% & 49.2\% & 36.8\% & 2.4\% & 1.16  & 2.8\% & 97.2\% \\
  3 & 1.05  & --    & 61.3\% & 36.3\% & 2.4\% & 0.41  & 3.2\% & 96.8\% \\
  4 & 0.54  & --    & 62.4\% & 35.2\% & 2.4\% & 0.15  & --   & 100\%  \\
  5 & 0.28  & --    & 61.6\% & 36.0\% & 2.4\% & 0.05  & --   & 100\%  \\
  6 & 0.14  & --    & --    & 97.6\% & 2.4\% & 0.02  & --   & 100\%  \\
  7 & 0.06  & --    & --    & 97.6\% & 2.4\% & 0.004 & --   & 100\%  \\
  8 & 0.03  & --    & --    & 97.6\% & 2.4\% & 0.001 & --   & 100\%  \\
  9 & 0.01  & --    & --    & 97.6\% & 2.4\% & 0.000 & --   & 100\%  \\
  \bottomrule
  \end{tabular}
  \end{table}

  Table~\ref{tab:lutstats} reports what calibration actually chose, and the
  \texttt{SDXL} column tracks the routing rule of Section~\ref{subsec:rthresh}
  exactly. \texttt{INT8} appears only in buckets $0$--$2$, precisely the buckets
  whose mean $\mathrm{SNR}(t)$ exceeds the $\tau_\mathrm{high} = 2$ gate, and
  disappears at $b = 3$ where $\mathrm{SNR}$ falls to $1.05$. \texttt{E5M2}
  saturates at $97.6\%$ from $b = 6$ onward, precisely where
  $\mathrm{SNR} < \tau_\mathrm{low} = 0.2$ forces Phase 1. \texttt{MXFP8} holds a
  constant $2.4\%$ in every bucket: this is $9/372$, the nine shortcut
  convolutions pinned by topology, and its flatness across $b$ is the signature of
  a rule that never consults a statistic. The three temporal regimes visible here
  ($b \le 2$, $3 \le b \le 5$, $b \ge 6$) are also why the bucket sweep is flat:
  $B = 5$ already resolves them.

  The \texttt{PixArt}-$\mathtt{\alpha}$ column is markedly less diverse. Its linear $\beta$ schedule
  decays $\mathrm{SNR}$ roughly two orders of magnitude faster, so the
  $\tau_\mathrm{low}$ gate fires from $b = 4$ onward rather than $b = 6$, and the
  remaining buckets are pushed to \texttt{E5M2} by kurtosis; the resulting
  \texttt{LUT} is $97$--$100\%$ \texttt{E5M2} throughout, effectively uniform. We
  state this plainly: on \texttt{PixArt}-$\mathtt{\alpha}$ the temporal component of the routing is
  close to degenerate, and the gains reported in
  Section~\ref{subsec:mainResults} come from the weight side -- group-wise \texttt{INT8-asym} at $g = 32$ for 8 bits and the joint $(g, f)$ search at 4 bits -- together with the dynamic
  per-sample scaling isolated in Table~\ref{tab:ab_fork}. The routing machinery is
  still correct on this backbone, but its degrees of freedom are not exercised;
  the same fixed thresholds behave differently only because $\mathrm{SNR}(t)$ is a
  property of the schedule, not of the method.

\subsection{Qualitative Samples}
\label{app:qualitative}

\begin{figure}[htbp]
    \centering
    \begin{tabular}{ ||>{\columncolor{Salmon!25}}c:>{\columncolor{TealBlue!25}}c:>{\columncolor{Goldenrod!25}}c|| }
        
        \hline
        \textbf{UNet Diffusion} & \textbf{Few-Step Distilled} & \textbf{Diffusion Transformer} \\
        \hline\hline

        \rowcolor{Gray!15} 
        \multicolumn{3}{||c||}{Half-Precision Float Baseline: \texttt{FP16 W$_{16}$A$_{16}$}} \\
        \hline
         & & \\[5pt]
         
        \fourblock{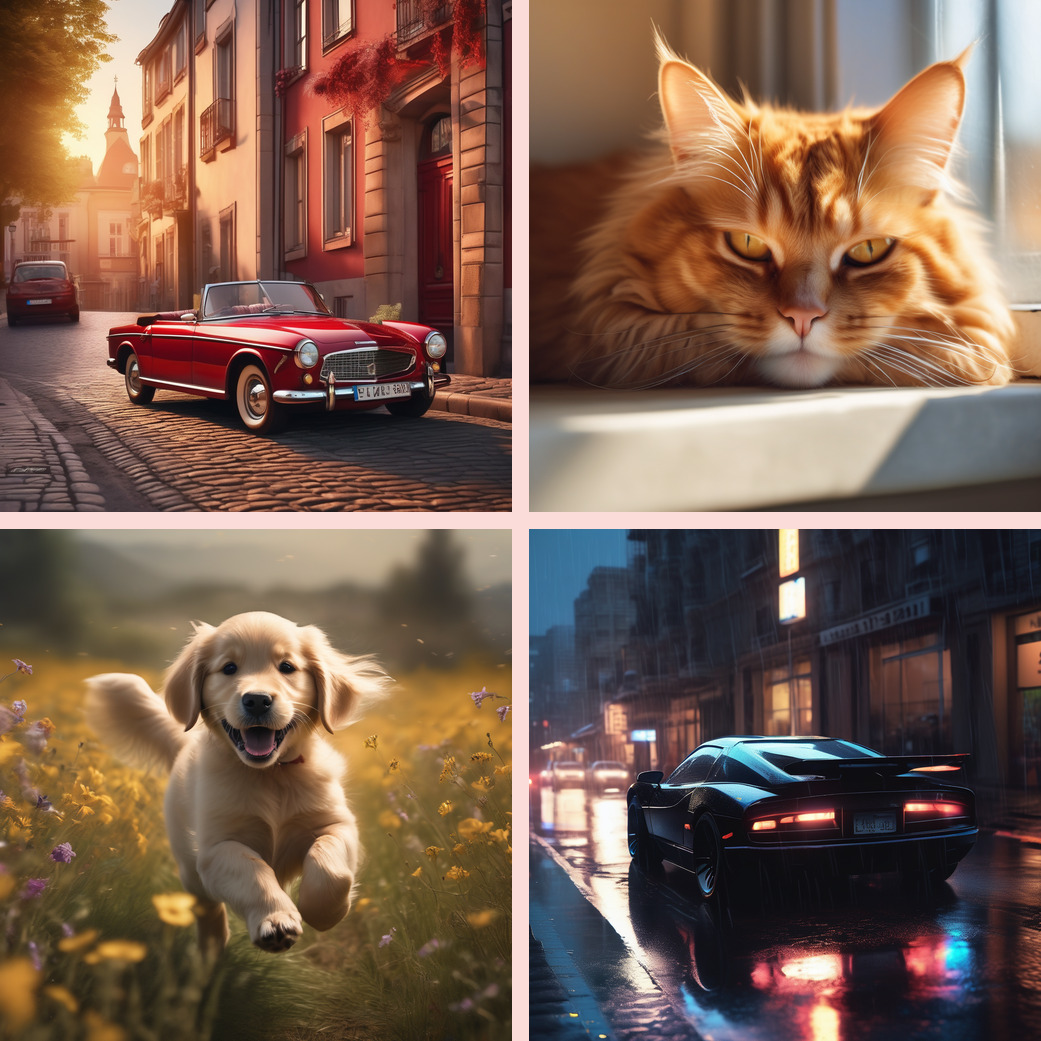}{SDXL} & 
        \fourblock{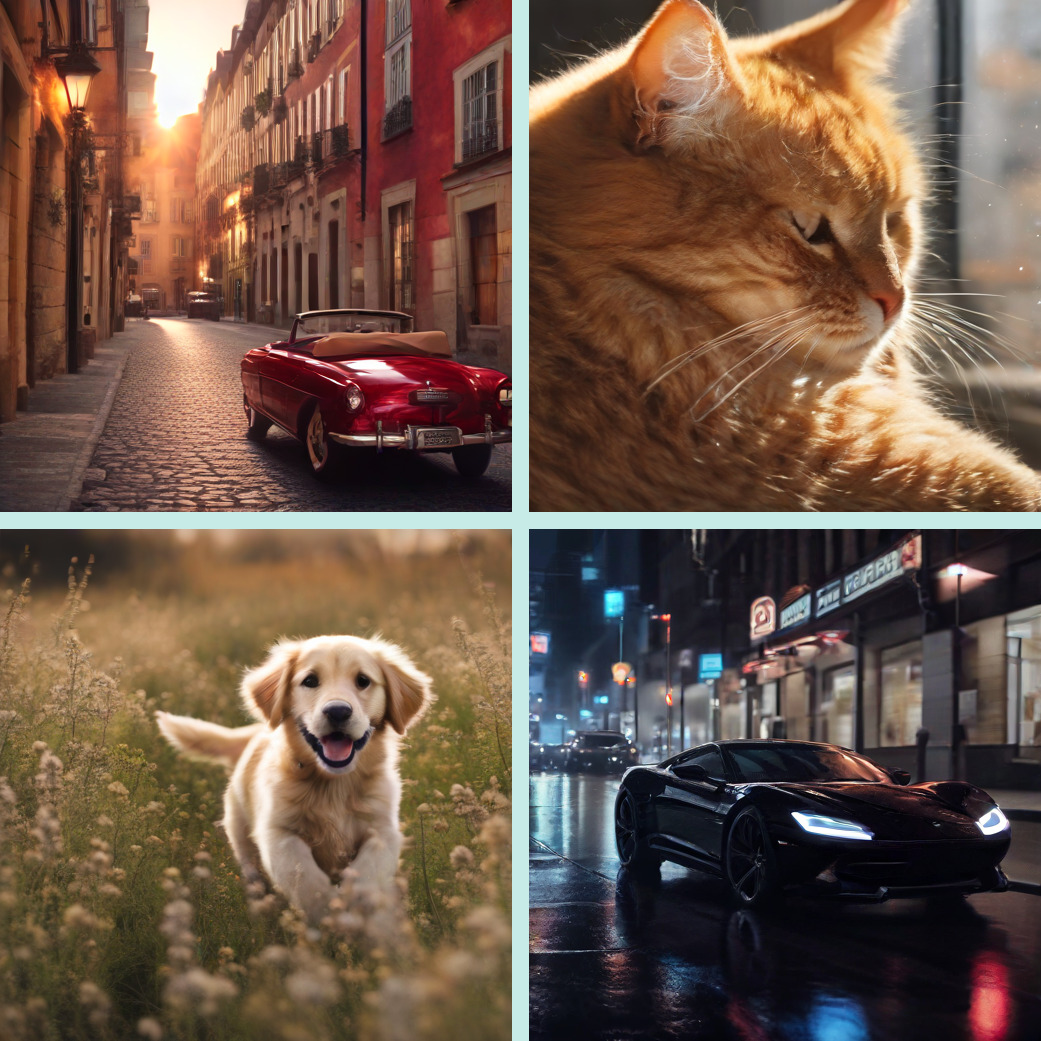}{SDXL-Turbo} & 
        \fourblock{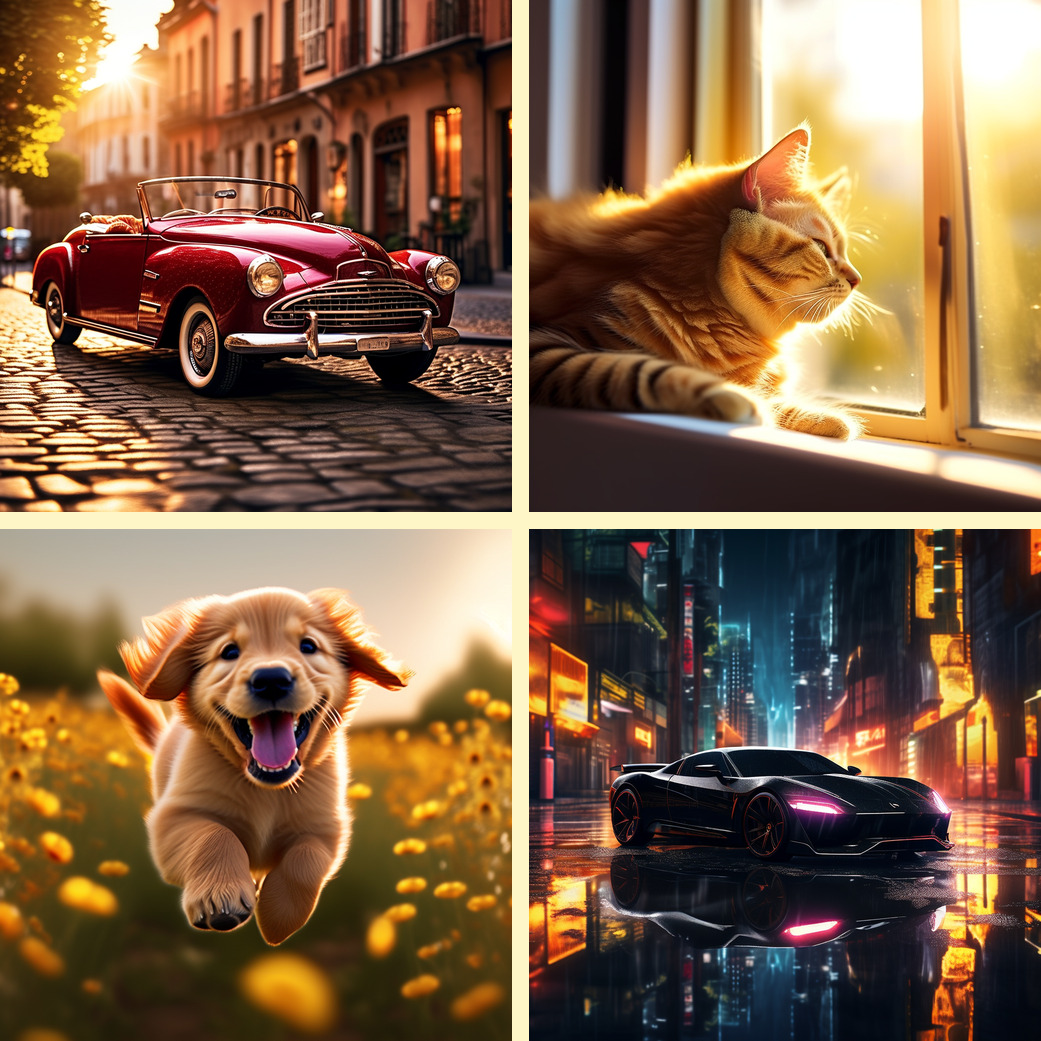}{PixArt-$\mathtt{\alpha}$} \\[5pt] 

        \hline
        \rowcolor{Gray!15} 
        \multicolumn{3}{||c||}{State-of-the-Art Quantized Baselines: \texttt{W$_{4}$A$_{8}$}} \\
        \hline
         & & \\[5pt]
        
        \fourblock{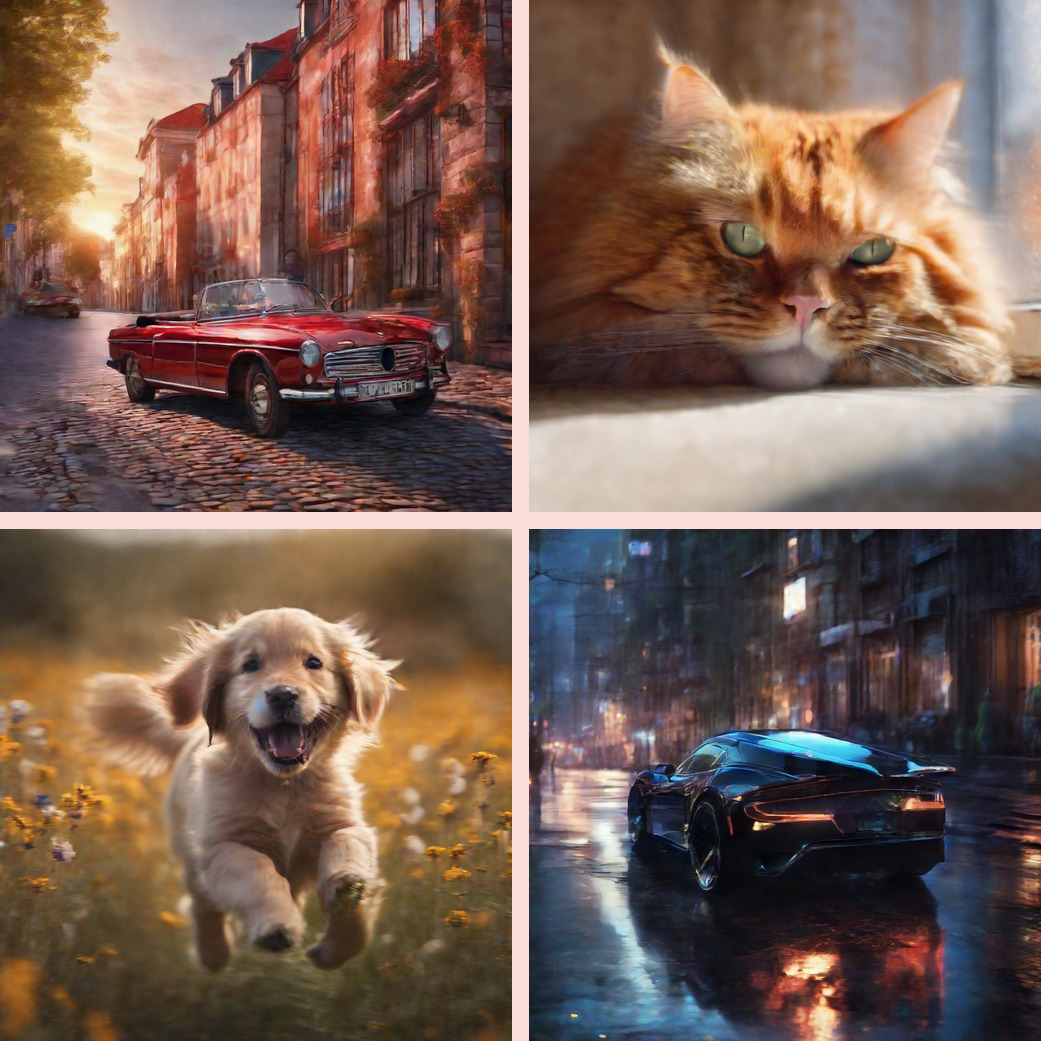}{Q-Diffusion} & 
        \fourblock{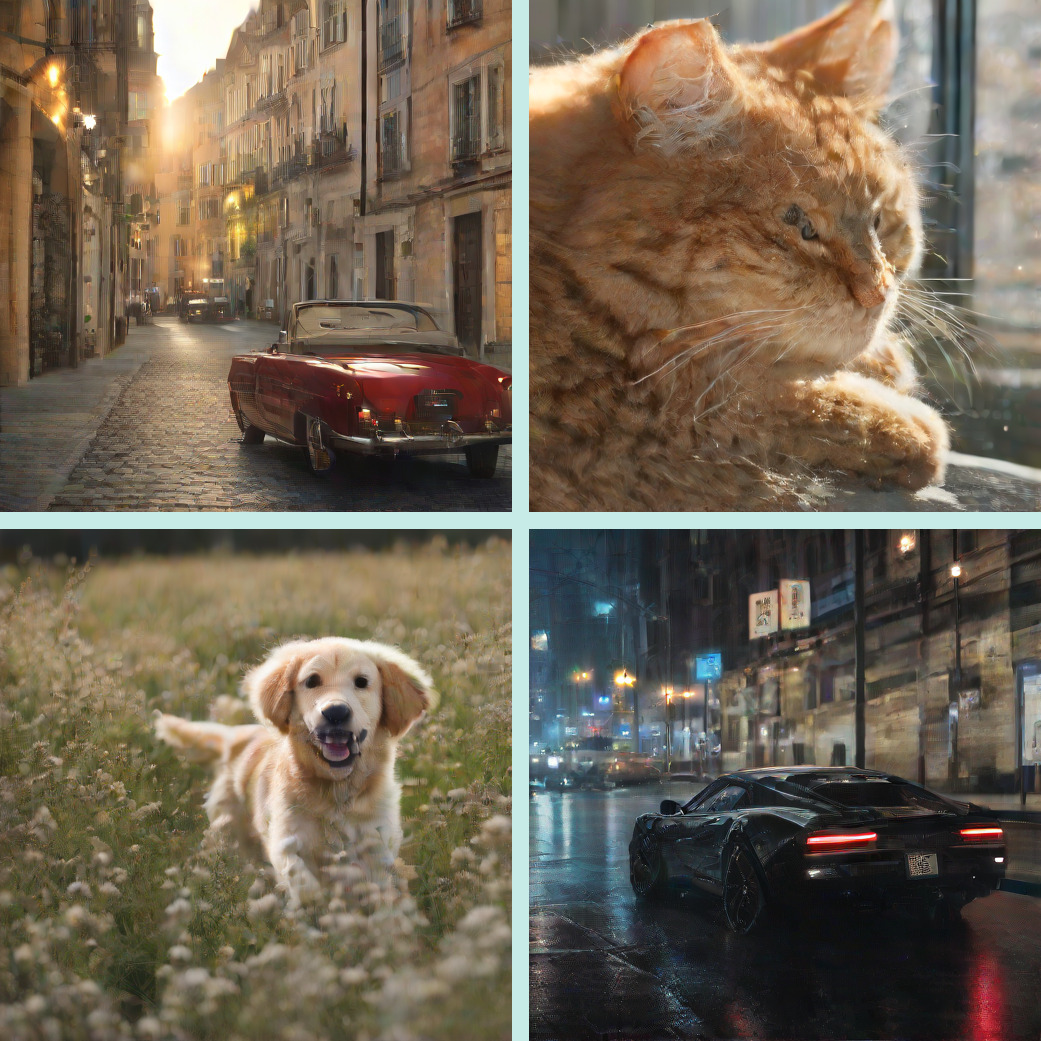}{MixDQ} & 
        \fourblock{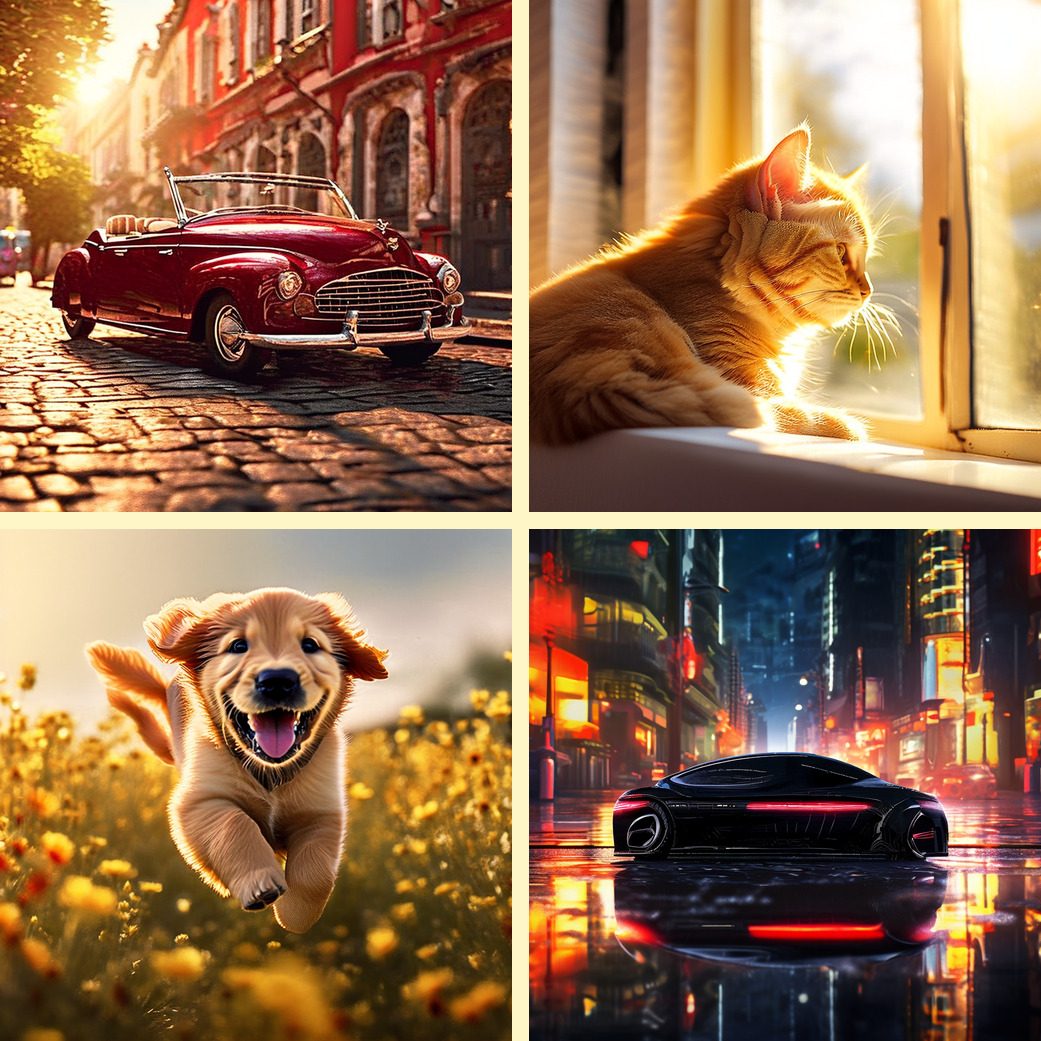}{Q-DiT} \\[5pt]

        \hline
        \rowcolor{Gray!15} 
        \multicolumn{3}{||c||}{Ours: \texttt{Chameleon W$_{4}$A$_{8}$}} \\
        \hline
         & & \\[5pt]
        
        \fourblock{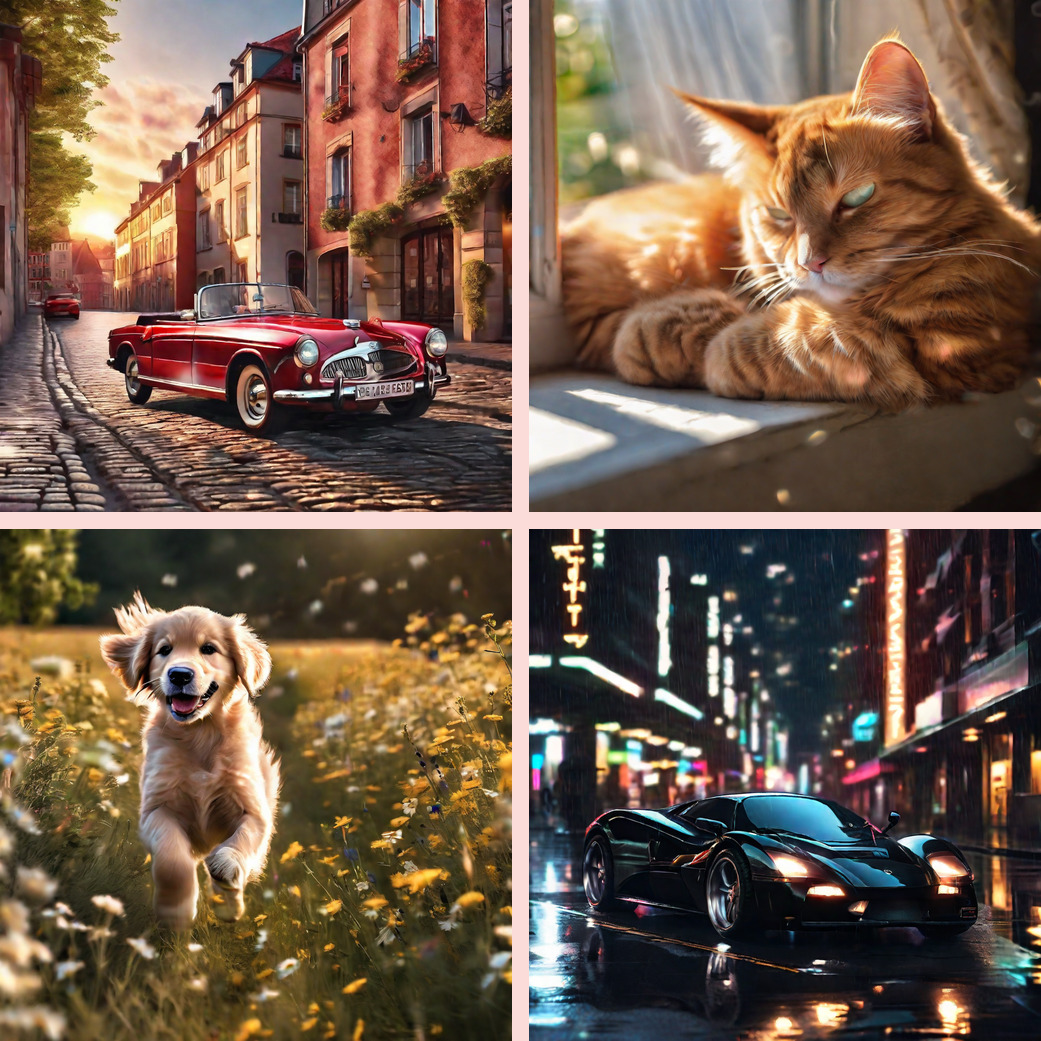}{Chameleon} & 
        \fourblock{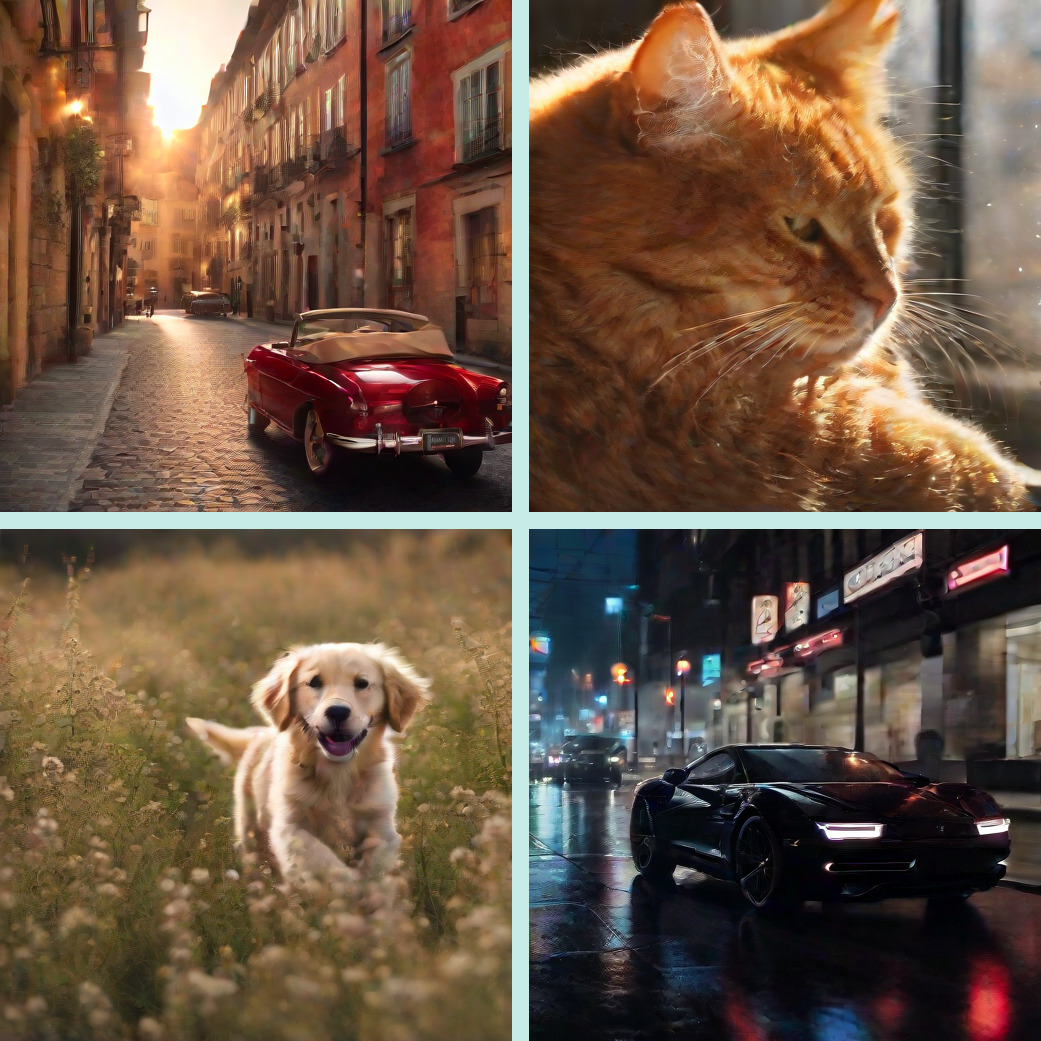}{Chameleon} & 
        \fourblock{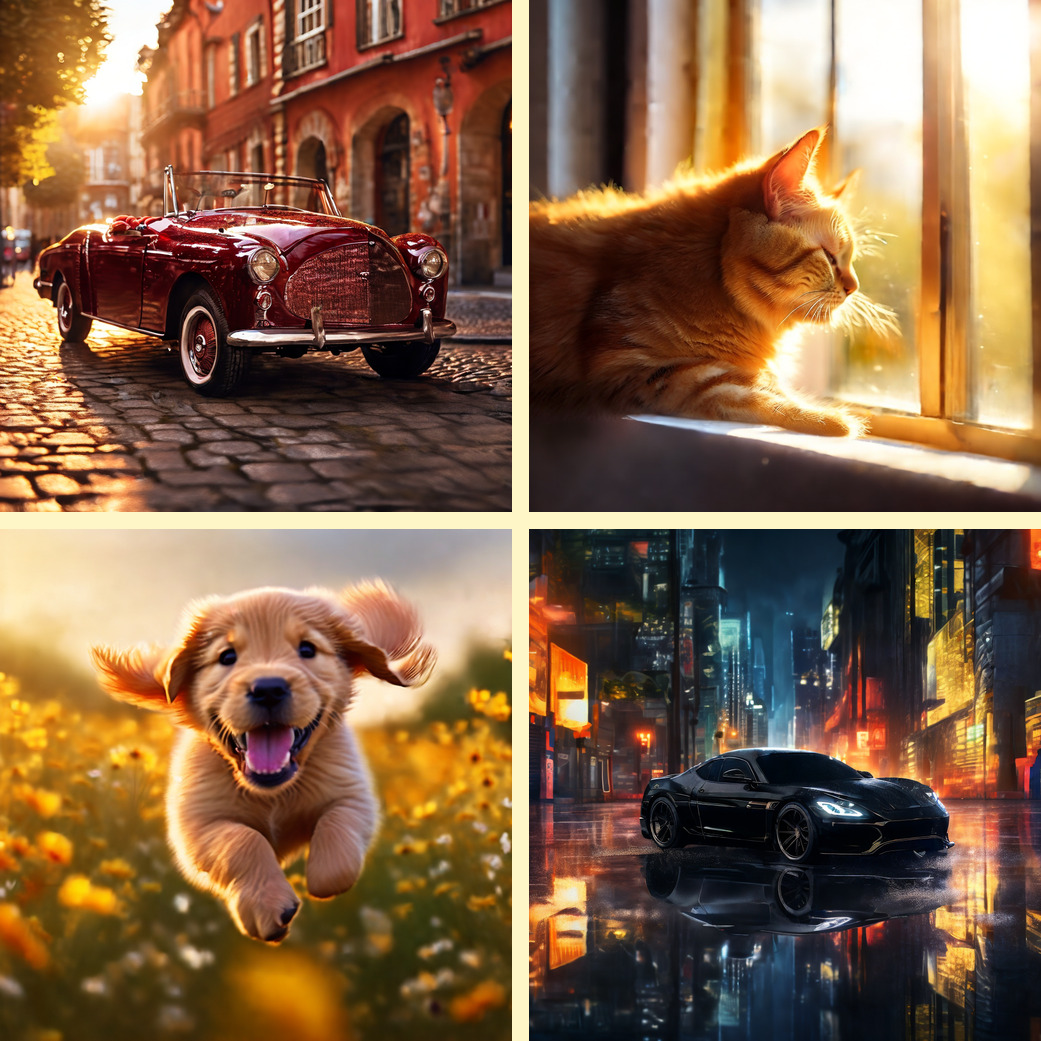}{Chameleon} \\
        
        \hline
        
    \end{tabular}
    
    \caption{Qualitative comparison on four prompts at W$_{4}$A$_{8}$. Rows (top to bottom): \texttt{FP16} reference (\texttt{SDXL}, \texttt{SDXL-Turbo}, \texttt{PixArt}-$\mathtt{\alpha}$); per-architecture \texttt{PTQ} baseline (\texttt{Q-Diffusion}, \texttt{MixDQ}, \texttt{Q-DiT} respectively); Chameleon. Columns (left to right): model architectures (UNet Diffusion, Few-Step Distilled, Diffusion Transformer). Within each $2 \times 2$ block, the four prompts are, in reading order, the {\color{Bittersweet} \textbf{cherry-red convertible}}, the {\color{BurntOrange}\textbf{orange tabby cat}}, the {\color{ForestGreen}\textbf{golden retriever puppy}}, and the \textbf{black sports car}; the prompts are given verbatim in the released code. Seeds are fixed across rows so that the same prompt index produces the same latent initialization under every quantization configuration.}
    \label{fig:36images}
\end{figure}

Figure~\ref{fig:36images} shows four prompts per backbone for the \texttt{FP16} reference, the state-of-the-art \texttt{PTQ} baseline, and Chameleon at W$_{4}$A$_{8}$; Table~\ref{tab:sharpness} quantifies the high-frequency detail visible in it.

\begin{table}[ht]
\centering
\caption{High-frequency detail retention, measured as the Variance of the Laplacian (\texttt{VL}), a standard image sharpness and focus measure \citep{pechpacheco2000diatom,pertuz2013focus}, over 400 images per configuration. Images are matched by filename across configurations, so every row compares the same prompts under the same seeds. Higher values indicate more high-frequency content; the \texttt{FP16} column is the unquantized reference for that backbone, not a target to exceed.}
\label{tab:sharpness}
\begin{tabular}{@{}llccc@{}}
\toprule
\textbf{Family} & \textbf{Backbone} & \textbf{\texttt{FP16}} &
\textbf{\texttt{PTQ} baseline} & \textbf{Chameleon} \\
\midrule
\texttt{UNet} diffusion & \texttt{SDXL}   & 522 & 605 & \textbf{662} \\
Few-step distilled       & \texttt{SDXL-Turbo} & 439 & 605 & \textbf{725} \\
Diffusion transformer    & \texttt{PixArt}-$\mathtt{\alpha}$ & 1287 & 1030 & \textbf{1057} \\
\bottomrule
\end{tabular}
\end{table}

As visible across the grid, on all model architectures, Chameleon retains more high-frequency detail than the competing \texttt{PTQ} baseline. Table~\ref{tab:sharpness} quantifies this over 400
prompt-matched images per configuration using the \texttt{VL}\footnote{We compute the measure as the variance of the $3\times3$ discrete Laplacian of the grayscale image (\texttt{scipy.ndimage.laplace}), which is numerically equivalent to the widely used \texttt{OpenCV} formulation with default kernel size.}: Chameleon reaches $662$ against \texttt{Q-Diffusion}'s $605$ on \texttt{SDXL}, $725$ against \texttt{MixDQ}'s $605$ on \texttt{SDXL-Turbo}, and $1057$ against \texttt{Q-DiT}'s $1030$ on \texttt{PixArt}-$\mathtt{\alpha}$.

\subsection{Number-Format Details}
\label{app:formats}

This appendix expands Section~\ref{subsec:formatp} of the main paper by
walking through each candidate format in Chameleon's palette and identifying the
kurtosis and \texttt{SNR} regimes (Appendix~\ref{app:phases}) in which \rev{the routing rule selects each, and why}.

\paragraph{\texttt{INT8} (asymmetric):} Encodes a tensor as 256 evenly spaced
bins between $\min$ and $\max$. Bin width $s = (\max - \min)/255$. Optimal under
a uniform prior; degrades quickly under heavy tails because $s$ inflates with the
outlier magnitude.

\paragraph{\texttt{FP8 E4M3}:} One sign bit, four exponent bits with bias 7,
three mantissa bits. Representable range $\pm 448$, smallest normal $2^{-6}$.
Mantissa precision is $2^{-3}$, so it cleanly resolves moderate dynamic range
with reasonable bin density.

\paragraph{\texttt{FP8 E5M2}:} One sign, five exponent, two mantissa.
Representable range $\pm 57344$, smallest normal $2^{-14}$. The five exponent
bits give it nine orders of magnitude of dynamic range. \rev{At the shipped calibrated scale, however, it is not more accurate than \texttt{E4M3} on real \texttt{SDXL} activations (relative error $4.5\%$ vs $2.3\%$). About half of its benefit in Phase 1 behaves like a regularizer: replacing the \texttt{E5M2}-routed slots with \texttt{E4M3} at the same range raises \texttt{FID} from $19.35$ to $20.13$, and \texttt{E4M3} plus zero-mean noise of \texttt{E5M2}'s error energy recovers about half of that gap ($19.74$; Table~\ref{tab:reb_e5m2}).}

\paragraph{\texttt{MXINT8} / \texttt{MXFP8}:} A microscaling block of 32 elements
carries one shared 8-bit exponent (\texttt{E8M0}); the elements themselves are
stored in \texttt{INT8} or \texttt{FP8} respectively. The shared exponent is
computed per block as
$E_\mathrm{block} = \lfloor \log_2 \max_i |x_i| \rfloor$, so a tensor containing
both $10^{-3}$ and $10^{2}$ values can be represented without either being
crushed: the relevant block selects its own exponent.
Appendix~\ref{app:mxconformance} records where our implementation conforms to the
\texttt{OCP} specification and what conformance costs. This is exactly the medicine for the wide-tailed concatenations at \texttt{UNet} shortcuts ($58\times$ median, up to $163\times$ peak-to-median channel range; Figure~\ref{fig:unet}).

\paragraph{\texttt{INT4} / \texttt{NF4} / \texttt{FP4} / \texttt{MXINT4} /
\texttt{MXFP4}:} At 4 bits, the choice of code matters even more. \texttt{INT4} is
uniform, in symmetric and asymmetric variants; \texttt{NF4}
\citep{dettmers2023qlora} is the quantile-optimal code for a unit Gaussian;
\texttt{FP4 E2M1} has a bias toward small values that tracks ReLU-like sparsity (one sign, two exponent and one mantissa bit, with representable values $\{0, \pm0.5, \pm1, \pm1.5, \pm2, \pm3, \pm4, \pm6\}$);
and the two \texttt{MX} variants add a shared block exponent on top of a narrow
integer or floating-point body. All six are available to the per-output-channel
search on \texttt{SDXL}; the few-step fork searches all but \texttt{INT4-sym} per layer jointly with the group size, and the \texttt{DiT} fork searches the
non-\texttt{MX} subset
$\{\texttt{INT4-asym}, \texttt{NF4}, \texttt{FP4 E2M1}\}$ jointly with the group
size (Section~\ref{subsec:archFork}), since an explicitly searched group size
already supplies the block granularity the \texttt{MX} formats would otherwise
contribute.

\rev{The kurtosis gates mark where \texttt{INT8} stops preserving the bulk of the distribution. Under a generalized-Gaussian model with the shipped static scales, energy-weighted \texttt{MSQE} is dominated by a few outliers (its \texttt{INT8}/\texttt{E4M3} crossover lies at $\kappa \approx 6$--$9$), whereas bulk \texttt{SQNR} ($|x| \le$ 99th percentile) crosses at $\kappa \approx 5$; $\kappa_\mathrm{high} = 5$ marks that crossover and $\kappa_\mathrm{low} = 3$ is a conservative bound below it (Appendix~\ref{app:rebuttal}).} The \texttt{SNR} gates $\{0.2, 2.0\}$ do not come from that model: they are
read off the noise schedule, and mark where the schedule enters and leaves its
noise-dominated phase.

\paragraph{Why fixed thresholds survive different noise schedules:}
The backbones we evaluate use different $\beta$ schedules -- scaled-linear for
\texttt{SDXL} and \texttt{SDXL-Turbo}, linear for \texttt{PixArt}-$\mathtt{\alpha}$ -- so
$\mathrm{SNR}(t)$ spans $\sim\!10^{3}$ down to $\sim\!10^{-2}$ on the former and
$\sim\!10^{4}$ down to $\sim\!10^{-4}$ on the latter. Applying the \emph{same}
gates $(\tau_\mathrm{low}, \tau_\mathrm{high}) = (0.2, 2.0)$ therefore partitions
the schedule differently: four of ten buckets fall into Phase 1 on \texttt{SDXL}
against six of ten on \texttt{PixArt}-$\mathtt{\alpha}$. This is the point rather than a
complication. The thresholds are stated in terms of $\mathrm{SNR}(t)$, which each
model's own noise schedule supplies, so they adapt to the schedule without being
retuned -- which is what lets the hyperparameter footprint stay fixed across
backbones (Section~\ref{subsec:calibCost}). The prediction is exact: the buckets
the gate marks as Phase 1 on \texttt{PixArt}-$\mathtt{\alpha}$ ($b \ge 4$) are precisely those the
shipped \texttt{LUT} fills with $100\%$ \texttt{FP8 E5M2}
(Appendix~\ref{app:lutstats}).


\subsection{MX Conformance and Its Cost}
\label{app:mxconformance}

The \texttt{OCP} \texttt{MX} specification requires the per-block shared scale to
be a power of two -- an \texttt{E8M0} exponent -- rather than an unconstrained
floating-point value. Our activation path implements this exactly: the block
scale is $2^{\lfloor \log_2 \max_i |x_i| \rfloor}$, so normalized values land in
$(-2, 2)$ and the simulation matches what conforming \texttt{MX} hardware
computes. The \texttt{SDXL} weight-side implementation deviates, using an
unconstrained per-block scale, which is strictly more expressive than the
specification allows; the few-step path's \texttt{MX} weights already use
power-of-two scales, and the \texttt{DiT} path selects no \texttt{MX} weight
format at all.

We measured the cost of conformance directly by rounding the weight-side scale
to the nearest power of two on real \texttt{SDXL} weights. It is $3.5$\,dB of
\texttt{SQNR} -- $45.4 \rightarrow 41.9$ for \texttt{MXINT8} and
$20.3 \rightarrow 16.7$ for \texttt{MXINT4} -- close to the theoretical worst
case of $6$\,dB from forfeiting up to one bit of range. The consequence for the
palette is larger than that figure suggests. Under an unconstrained scale,
block-scaled \texttt{MXINT8} wins essentially every output channel at 8 bits,
which would make the weight palette degenerate; under \texttt{E8M0} the
per-channel selection of Section~\ref{subsec:weightSel} distributes across
\texttt{INT8-asym} ($59\%$), \texttt{MXINT8} ($37\%$) and \texttt{INT8-sym}
($4\%$). Conformance is therefore what makes the per-channel search meaningful at
8 bits rather than a formality.

The \texttt{SDXL} weight-side numbers reported in this paper correspond to the
unconstrained variant and should be read as an upper bound on what a conforming \texttt{MX}
weight kernel would deliver; the activation-side numbers are conformant as
reported.

\subsection{Extended Discussion and Limitations}
\label{app:discussion}

\paragraph{Real hardware versus fake-quant:} All numbers reported in
Section~\ref{sec:experiments} use fake-quantization -- round-trip casts inside
\texttt{FP16} -- because real
\texttt{INT4}/\texttt{NF4}/\texttt{FP4}/\texttt{MX4} kernels are not yet
uniformly available across the three backbones we evaluate. \rev{Appendix~\ref{app:rebuttal} adds first real-kernel measurements: an exact Chameleon kernel on Ada ($1.09\times$ end to end, $3\times$ smaller weights) and a hardware-native \texttt{MX} variant on Blackwell ($1.74\times$ faster GEMMs, end-to-end parity because attention dominates). Quantizing attention and extending the kernels to Hopper are left to future work.}

\paragraph{Composition with sampler adaptation:} \texttt{Q-Sched}
\citep{frumkin2025qsched} keeps the quantized weights frozen and learns two
scalar preconditioning coefficients per sampler step that warp the integration
trajectory. Their motivation -- that aggressive quantization changes the
underlying vector field, so the sampler optimized for \texttt{FP16} is no longer
optimal -- is orthogonal to Chameleon's \rev{format choice}. Composing the
two should be straightforward: run Chameleon's calibration first to fix the
format \texttt{LUT}, then run \texttt{Q-Sched}'s preconditioning search on top of
the now-quantized weights. We have not measured this composition and leave it to
future work.

\paragraph{Image scope:} The current evaluation is limited to image generation.
We expect the same routing principle to carry over to video diffusion
transformers -- the macro physics of the diffusion schedule, and therefore the
kurtosis/\texttt{SNR} phase structure that Chameleon exploits, is unchanged in
the temporal setting -- but the temporal-attention layers in video \texttt{DiT}s
introduce additional sensitivity that we have not yet characterized. Extending
Chameleon to video \texttt{DiT}s is a natural next step.

\paragraph{Sub-4-bit territory:} We do not address W$_{2}$ or W$_{1}$
quantization. At those bit-widths the discrete approximation error overwhelms
what any post-training routing scheme can recover, and quantization-aware
training with techniques such as \texttt{BinaryDM}'s evolvable-basis binarizer
\citep{zheng2024binarydm} becomes necessary. Chameleon's calibration pass could
in principle initialize such a \texttt{QAT} phase, but doing so is outside the
scope of a \texttt{PTQ} paper.

\paragraph{Large per-channel format metadata:} At extreme model scale (e.g.\ a
hypothetical 20-billion-parameter image \texttt{DiT}) the per-output-channel
format byte starts to add measurable overhead. We expect that a per-layer format
choice with per-channel scale would recover most of the benefit at a fraction of
the metadata cost, since within a single conv layer the dominant format choice
tends to be shared across most output channels.

\paragraph{Calibration on out-of-distribution prompts:} Our calibration uses 128
\texttt{COCO} captions for all three runs. We have not stress-tested
generalization to dramatically different prompt distributions -- for instance,
calibrating on \texttt{COCO} and evaluating on text-rendering or
hands-and-faces benchmarks. A practitioner deploying Chameleon on a
domain-specific generator should re-calibrate with a domain-matched prompt set.

{
\subsection{Additional Experiments (Rebuttal)}
\label{app:rebuttal}

\paragraph{Protocol:} To fit the rebuttal period, the experiments below use the first 5{,}000 \texttt{COCO} val2014 captions in the same order, with the same per-image seeds, clean-\texttt{FID} reference set and \texttt{CLIP-L/14} scoring as Table~\ref{tab:combined_results}. \texttt{FID} depends on sample size, so 5k numbers are comparable only with each other. Within each table every row ran on the same GPU type and software stack (NVIDIA GB200 unless marked A5000; images from different GPU types are not bit-identical, see the kernel paragraph below).

\paragraph{Threshold sensitivity:} One threshold changed at a time, \texttt{SDXL} W$_4$A$_8$ (Table~\ref{tab:reb_thr}). \texttt{FID} moves by at most $0.13$, comparable to the spread across calibration seeds ($0.05$--$0.14$, Table~\ref{tab:reb_seeds}).

\begin{table}[htbp]
\centering\small
\caption{Routing-threshold sensitivity, \texttt{SDXL} W$_4$A$_8$, 5k, GB200.}
\label{tab:reb_thr}
\begin{tabular}{@{}lcc@{}}
\toprule
Setting & \texttt{FID} $\downarrow$ & \texttt{CLIP} $\uparrow$ \\
\midrule
Shipped ($\kappa_\mathrm{low}{=}3$, $\kappa_\mathrm{high}{=}5$, $\tau{=}(0.2, 2)$) & 19.44 & 26.59 \\
$\kappa_\mathrm{low} = 2.5$ / $3.5$ & 19.40 / 19.42 & 26.59 / 26.58 \\
$\kappa_\mathrm{high} = 4.5$ / $6.0$ & 19.52 / 19.40 & 26.59 / 26.58 \\
$\tau = (0.1, 1)$ / $(0.5, 5)$ & 19.39 / 19.44 & 26.60 / 26.59 \\
\bottomrule
\end{tabular}
\end{table}

\paragraph{Basis of the kurtosis gates:} We re-derived the per-format errors under a generalized-Gaussian model using exactly the shipped static scaling rules. Energy-weighted \texttt{MSQE} is dominated by a handful of outliers; under it the \texttt{INT8}/\texttt{E4M3} crossover lies at $\kappa \approx 6$--$9$ depending on the calibration sample count, so it is not the basis of the gates. What matters for generation is how well a format preserves the bulk of the distribution ($|x| \le$ 99th percentile). On that measure \texttt{INT8} leads \texttt{FP8 E4M3} at $\kappa = 4.5$ ($34.2$ vs $31.4$\,dB bulk \texttt{SQNR}) and trails it at $\kappa = 5.3$ ($31.2$ vs $31.6$\,dB), and the share of nonzero activations that \texttt{INT8} rounds to zero grows from $1.7\%$ ($\kappa = 3$) to $6.4\%$ ($\kappa = 6$) and $35\%$ ($\kappa = 25$). $\kappa_\mathrm{high} = 5$ marks that crossover, and $\kappa_\mathrm{low} = 3$ is a conservative bound below it.

\paragraph{Error norm for weight-format selection:} Formats ranked by L1, L$_\infty$ or cosine distance instead of squared L2, with the \texttt{FP16} fallback held fixed (Table~\ref{tab:reb_norm}). L1, L2 and cosine agree within $0.15$ (GB200) and $0.3$ (A5000) \texttt{FID}; only L$_\infty$, which ranks formats by their single worst element, is consistently worse.

\begin{table}[htbp]
\centering\small
\caption{Weight-format ranking objective, \texttt{SDXL} W$_4$A$_8$, 5k.}
\label{tab:reb_norm}
\begin{tabular}{@{}lccc@{}}
\toprule
Objective & \texttt{FID} (GB200) & \texttt{FID} (A5000) & \texttt{CLIP} (GB200) \\
\midrule
L2 (shipped) & 19.44 & 19.35 & 26.59 \\
L1 & 19.34 & 19.09 & 26.58 \\
cosine & 19.48 & 19.38 & 26.55 \\
L$_\infty$ & 19.97 & 19.75 & 26.46 \\
\bottomrule
\end{tabular}
\end{table}

\paragraph{Role of \texttt{FP8 E5M2}:} On real \texttt{SDXL} activations at the shipped calibrated range, \texttt{E5M2} has about twice the relative error of \texttt{E4M3} ($4.5\%$ vs $2.3\%$). Changing only the (layer, bucket) slots the rule routes to \texttt{E5M2} (Table~\ref{tab:reb_e5m2}) shows that the more accurate \texttt{E4M3} is worse, and that \texttt{E4M3} plus zero-mean multiplicative noise with exactly \texttt{E5M2}'s error energy recovers about half of the gap: about half of \texttt{E5M2}'s benefit behaves like a regularizer.

\begin{table}[htbp]
\centering\small
\caption{Format of the \texttt{E5M2}-routed slots, \texttt{SDXL} W$_4$A$_8$, 5k, A5000.}
\label{tab:reb_e5m2}
\begin{tabular}{@{}lcc@{}}
\toprule
\texttt{E5M2}-routed slots use & \texttt{FID} $\downarrow$ & \texttt{CLIP} $\uparrow$ \\
\midrule
\texttt{E5M2} (shipped rule) & \textbf{19.35} & 26.55 \\
\texttt{E4M3}, same calibrated range & 20.13 & 26.47 \\
\texttt{E4M3} + noise of \texttt{E5M2}'s error energy & 19.74 & 26.63 \\
\bottomrule
\end{tabular}
\end{table}

\paragraph{Comparison with \texttt{MoFQ}- and \texttt{TDQ}-style adaptation:} Table~\ref{tab:reb_mofq}. The \texttt{MoFQ}-style arms choose one format per layer by calibration error (activations: \texttt{INT8-asym} vs \texttt{FP8 E4M3}, which picks \texttt{INT8} for $86\%$ of layers; weights: \texttt{INT4-asym} vs \texttt{FP4 E2M1} per layer, which picks \texttt{FP4} for 370 of 372 layers, with Chameleon's \texttt{FP16} channel fallback unchanged). The \texttt{TDQ}-style arm is a training-free proxy of \citet{so2023tdq}: all non-shortcut activations in \texttt{INT8-asym}, one calibrated range per (layer, timestep bucket), $B = 10$. Calibrated per-bucket ranges are exceeded at inference and, without widening, generation collapses (as in the closely related \texttt{single\_int8} row of Table~\ref{tab:ab_palette}, \texttt{FID} $180.5$, which also forces the shortcut layers to \texttt{INT8}); widening each range by $1.5\times$ removes the collapse.

\begin{table}[htbp]
\centering\small
\caption{Format selection rule vs error-based and scale-only adaptation, \texttt{SDXL} W$_4$A$_8$, 5k, A5000.}
\label{tab:reb_mofq}
\begin{tabular}{@{}llcc@{}}
\toprule
Method & What adapts & \texttt{FID} $\downarrow$ & \texttt{CLIP} $\uparrow$ \\
\midrule
\texttt{TDQ}-style (all \texttt{INT8}, per-bucket ranges, $1.5\times$) & scale per timestep & 23.92 & 26.18 \\
\texttt{MoFQ}-style, activations & format per layer, by error & 20.14 & 26.47 \\
\texttt{MoFQ}-style, weights + activations & format per layer, by error & 20.06 & 26.50 \\
Chameleon & format, $\kappa$/\texttt{SNR} rule & \textbf{19.35} & \textbf{26.55} \\
\bottomrule
\end{tabular}
\end{table}

\paragraph{Timestep buckets:} Table~\ref{tab:reb_seeds} repeats the bucket ablation with three calibration seeds. At A$_8$ one bucket is as good as or better than ten. At 6-bit activations (\texttt{INT6-asym}, \texttt{FP6 E2M3}, \texttt{FP6 E3M2} routed by the same rule, \texttt{MXFP6} on shortcut layers), per-layer routing beats the best single format by $0.3$--$0.6$ \texttt{FID} and five buckets give the best result; an A100 replicate gives $B{=}1$ 19.50 and $B{=}10$ 19.66. On \texttt{PixArt}-$\alpha$ W$_4$A$_6$, $B{=}10$ and $B{=}1$ give 28.94 and 29.02 (the $B{=}10$ setting is $+0.10$ relative to W$_4$A$_8$).

\begin{table}[htbp]
\centering\small
\caption{Timestep buckets, \texttt{SDXL}, 5k, GB200. Top: W$_4$A$_8$, three calibration seeds. Bottom: W$_4$A$_6$.}
\label{tab:reb_seeds}
\begin{tabular}{@{}lccccc@{}}
\toprule
W$_4$A$_8$ & Seed 0 & Seed 1000 & Seed 2000 & Mean $\pm$ SD & \texttt{CLIP} \\
\midrule
$B = 1$ & 19.18 & 19.25 & 19.28 & \textbf{19.24} $\pm$ 0.05 & 26.71 \\
$B = 10$ & 19.44 & 19.37 & 19.64 & 19.48 $\pm$ 0.14 & 26.57 \\
\midrule
W$_4$A$_6$ & & & & \texttt{FID} & \texttt{CLIP} \\
\midrule
Single format (\texttt{FP6 E2M3}) & & & & 20.03 & 26.64 \\
Rule, $B = 1$ & & & & 19.62 & 26.84 \\
Rule, $B = 5$ & & & & \textbf{19.40} & 26.73 \\
Rule, $B = 10$ & & & & 19.70 & 26.71 \\
\bottomrule
\end{tabular}
\end{table}

\paragraph{Group size versus format on \texttt{PixArt}-$\alpha$:} Table~\ref{tab:reb_group}. At equal $g = 32$ Chameleon is $1.2$ \texttt{FID} better than \texttt{Q-DiT}: about $0.9$ from activation routing (\texttt{Q-DiT} vs Chameleon \texttt{INT4}-only) and $0.3$ from adaptive weight formats. Reducing \texttt{Q-DiT}'s group size from 128 to 32 makes it $2.7$ \texttt{FID} worse. At W$_8$A$_8$, \texttt{Q-DiT} gives 32.95 ($g{=}128$) and 32.74 ($g{=}32$) against 29.48 for Chameleon.

\begin{table}[htbp]
\centering\small
\caption{Equal group size ($g = 32$): weight format and activation routing, \texttt{PixArt}-$\alpha$ W$_4$A$_8$, 5k, GB200.}
\label{tab:reb_group}
\begin{tabular}{@{}lcllcc@{}}
\toprule
Method & $g$ & Weight formats & Activations & \texttt{FID} $\downarrow$ & \texttt{CLIP} $\uparrow$ \\
\midrule
\texttt{Q-DiT} & 32 & \texttt{INT4} & \texttt{INT8} dynamic & 30.03 & 25.86 \\
Chameleon & 32 & \texttt{INT4} only & routed \texttt{FP8} & 29.17 & 25.66 \\
Chameleon & 32 & \texttt{FP4} only & routed \texttt{FP8} & 29.88 & 25.53 \\
Chameleon & 32 & \texttt{NF4} only & routed \texttt{FP8} & 29.92 & 25.68 \\
Chameleon & 32 & adaptive & routed \texttt{FP8} & 28.84 & 25.68 \\
\bottomrule
\end{tabular}
\end{table}

\paragraph{\texttt{ViDiT-Q}, \texttt{FP4DiT} and \texttt{PTQ4DiT}:} \texttt{ViDiT-Q} \citep{zhao2025viditq} was run in its own \texttt{PixArt}-$\alpha$ pipeline (its own \texttt{FP16} reference and DPM-Solver sampler) with our prompts, seeds and scoring. Its released \texttt{PixArt} ``w4a8'' configurations use 6-bit weights with per-channel scales, and per-channel W$_4$ collapses in its pipeline (\texttt{FID} 140.2), so we added group-wise ($g = 128$) W$_4$ scales to its quantizer, matching \texttt{Q-DiT}. Because the two pipelines have different \texttt{FP16} references, Table~\ref{tab:reb_vidit} compares each method's change relative to its own reference. \texttt{FP4DiT} \citep{chen2025fp4dit} evaluates \texttt{PixArt}-$\alpha$ at $512\times512$ on 10k \texttt{COCO}-2014 prompts in its own pipeline; its published table gives full precision 34.05, W$_4$A$_8$ 30.37 ($\Delta\,{-3.68}$) and W$_4$A$_6$ 40.61 ($\Delta\,{+6.56}$), with \texttt{Q-DiT} at 36.17 / 43.91 and \texttt{ViDiT-Q} at 39.11 / 56.54. Chameleon's W$_4$A$_6$ costs only $+0.10$ \texttt{FID} over its W$_4$A$_8$ under our protocol, against $+10.2$ for \texttt{FP4DiT}; these are published numbers at a different resolution, not reruns in our harness. \texttt{PTQ4DiT} \citep{wu2024ptq4dit} targets class-conditional \texttt{DiT-XL/2} on ImageNet and is outside our text-to-image setting.

\begin{table}[htbp]
\centering\small
\caption{\texttt{PixArt}-$\alpha$: \texttt{FID} and change relative to each pipeline's own \texttt{FP16} reference.}
\label{tab:reb_vidit}
\begin{tabular}{@{}lccc@{}}
\toprule
& \texttt{FP16} & W$_8$A$_8$ ($\Delta$) & W$_4$A$_8$ ($\Delta$) \\
\midrule
\texttt{ViDiT-Q} (own pipeline, 5k) & 31.01 & 30.18 ($-0.83$) & 27.11 ($-3.90$) \\
Chameleon (Table~\ref{tab:combined_results}) & 27.63 & 24.33 ($\mathbf{-3.30}$) & 22.24 ($\mathbf{-5.39}$) \\
\bottomrule
\end{tabular}
\end{table}

\paragraph{Real kernels:} On an RTX 6000 Ada (\texttt{FP8} tensor cores), a Triton kernel reproduces Chameleon's numerics exactly: weights are stored as 4-bit codes (\texttt{INT4-asym} and \texttt{FP4} decode exactly into \texttt{FP8} tensor-core operands; \texttt{NF4} uses \texttt{FP16} MMA), activations are per-tensor \texttt{FP8} with one round-to-nearest-even cast, and a single kernel serves every weight format through a 32-entry lookup table, so mixing formats across layers adds no kernel variants. Per-layer outputs match fake quantization within $9\times10^{-4}$ (Table~\ref{tab:reb_kernel}). On a GB200, a hardware-native variant with power-of-two block scales (\texttt{E8M0}, block 32) runs every format as one cuBLAS \texttt{MXFP8} GEMM, bit-exact against its own reference: GEMMs are $1.74\times$ faster than \texttt{FP16} (4.00 vs 6.95\,ms per step), while end to end it matches \texttt{FP16} (494 vs 493\,ms) because attention and elementwise work take $70\%$ of each step. The power-of-two scaling costs $+0.93$ \texttt{FID} relative to exact Chameleon and still beats \texttt{Q-DiT} at $g = 32$. \texttt{FID} reproduces exactly across A100 and GB200, although individual images differ across GPU types.

\begin{table}[htbp]
\centering\small
\caption{\texttt{PixArt}-$\alpha$ $1024^2$, 20 steps, batch 1, RTX 6000 Ada.}
\label{tab:reb_kernel}
\begin{tabular}{@{}lccc@{}}
\toprule
& ms / image & vs \texttt{FP16} & Transformer weights \\
\midrule
\texttt{FP16} (cuBLAS) & 2{,}776 & $1.00\times$ & 1{,}165\,MB \\
Fake quantization (paper) & 4{,}712 & $0.59\times$ & -- \\
Chameleon, real kernel & \textbf{2{,}555} & $\mathbf{1.09\times}$ & \textbf{392\,MB} \\
Quantized linear layers only (CUDA graphs) & 1{,}206 vs 1{,}432 & $1.19\times$ & \\
\bottomrule
\end{tabular}
\end{table}

\paragraph{\texttt{SDXL-Turbo} format statistics and layer sensitivity:} All 777 quantized \texttt{UNet} layers are 4-bit; the joint (format, group size) search selects \texttt{INT4-asym} at $g = 32$ for 769 layers ($99.0\%$) and \texttt{NF4} at $g = 32$ for 8 (7 in the up-blocks); \texttt{FP4}, \texttt{MXINT4} and \texttt{MXFP4} are never chosen. Table~\ref{tab:reb_turbo} replaces Chameleon's choice with per-output-channel symmetric \texttt{INT4} in one block group at a time, keeping every layer at 4 bits. The up-block and ResNet layers, which receive skip-concatenated inputs, collapse; \texttt{MixDQ} avoids this with mixed precision (its ``W4'' keeps 106 of 410 up-block layers at 8 bits), whereas Chameleon stays at 4 bits throughout.

\begin{table}[htbp]
\centering\small
\caption{\texttt{SDXL-Turbo} W$_4$A$_8$, one block group switched to per-channel \texttt{INT4}, 5k.}
\label{tab:reb_turbo}
\begin{tabular}{@{}lcc@{}}
\toprule
Block group & \texttt{FID} $\downarrow$ & $\Delta$ \\
\midrule
none (Chameleon) & 26.72 & -- \\
FFN & 26.45 & $-0.27$ \\
mid block & 26.85 & $+0.13$ \\
attention & 27.18 & $+0.46$ \\
down blocks & 27.26 & $+0.54$ \\
ResNet blocks & 225.76 & collapse \\
up blocks & 244.22 & collapse \\
\bottomrule
\end{tabular}
\end{table}
}

\newpage
\section{Statements}

\subsection{AI Use Statement}
\label{stat:ai}

In this work, we used generative AI tools to implement methods. We have not used generative AI tools to generate synthetic data sets, to develop the theoretical models or conceptual frameworks underlying the method, to propose or refine hypotheses, to design or give feedback on research methodology or experiments, to support qualitative or thematic data analysis, to interpret results, to clean or reformat datasets, or for translation. The remaining tasks with required disclosure -- formulating mathematical claims, providing critical ingredients for proving mathematical claims, and assisting in the writing of proofs -- are not applicable to this work, which contains no formal proofs. Additionally, we used generative AI tools to create and edit software code, to create and modify scientific figures, to draft and edit parts of this paper, to suggest a structure for the supplementary material, to identify relevant literature, to format references, and to search for information. We have reviewed all AI-assisted work.

To elaborate: the research contributions of this paper are the authors'. The central idea -- treating the number format as a discrete variable indexed by (layer, timestep bucket) -- the kurtosis/\texttt{SNR} routing rule, the architectural fork, the design of the experimental protocol and of the ablation suite, and the interpretation of every result reported here were carried out by the authors. Generative AI assistance was confined to implementation and exposition. On the implementation side, it was used to write and debug the ablation harness, the per-arm monkey-patch shims, the figure-generation scripts, and several corrections to the quantization code, including a bucket-count propagation bug and a caption-alignment bug in the \texttt{CLIP} scorer (Appendix~\ref{app:clip_scaling}). It was also used to compute the auxiliary measurements reported in Appendices~\ref{app:qualitative} and~\ref{app:ablations}, such as the Laplacian sharpness statistics and the \texttt{FID}--\texttt{CLIP} correlation; the authors specified what to measure and drew the conclusions from the output. Each AI-assisted code change was validated against generated artifacts before use -- for instance by byte-comparing images across ablation arms to confirm that a patch had a non-trivial effect -- and every claim in AI-drafted text was checked by the authors against the artifacts on disk, with a number of claims removed or weakened as a result. Literature identified with AI assistance was verified against primary sources. We take responsibility for the final content of this work, including all text, claims, and artifacts produced with the aid of generative AI.

\subsection{Ethics Statement}
\label{stat:ethics}

This work studies post-training quantization of existing, publicly released text-to-image diffusion models. It involves no human subjects, no personally identifying information, and no new data collection: all evaluation uses the public \texttt{COCO}-2014 validation set \citep{lin2014coco} and publicly released model checkpoints under their respective licenses.

Our method inherits the characteristics of the models it compresses. Text-to-image diffusion models are known to reproduce social biases present in their training data and can be used to generate misleading or harmful imagery. Quantization does not introduce these properties, but neither does it remove them, and we did not evaluate whether format routing shifts a model's behavior on safety-relevant prompts -- our evaluation measures distributional fidelity and prompt adherence only. A practitioner deploying a quantized model should not assume that safety evaluations performed on the full-precision model transfer unchanged.

The intended benefit of this work is to reduce the memory and compute required to run these models, which lowers the barrier to on-device inference and reduces the energy cost of deployment. We note the dual-use character of that benefit: the same reduction makes generative models easier to deploy for misuse, and easier to run outside the monitoring that a hosted service provides. We judge the efficiency and accessibility benefits to outweigh this risk for a method that adds no capability beyond what the underlying models already have.

\subsection{Reproducibility Statement}
\label{stat:repro}

We will release the framework code together with the per-model calibration artifacts -- weight-format configurations and activation \texttt{LUT}s for all three backbones at both bit-widths -- so that every number in Section~\ref{sec:experiments} can be reproduced without repeating calibration. Their layout and sizes are given in Table~\ref{tab:calib_cost}.

\paragraph{Environment:} All experiments run on a single NVIDIA A100 80\,GB (driver 580.159.03) under Python 3.12, PyTorch 2.11 with CUDA 13.0, \texttt{diffusers} 0.37.1, \texttt{transformers} 5.5.4, \texttt{clean-fid} 0.1.35, NumPy 2.4 and SciPy 1.17. The three backbones are public checkpoints used without fine-tuning: \texttt{SDXL} base~1.0, \texttt{SDXL-Turbo}, and \texttt{PixArt}-$\mathtt{\alpha}$ \texttt{XL/2 1024-MS}.

\paragraph{Prompts and seeds:} Evaluation uses the \texttt{COCO}-2014 validation set \citep{lin2014coco}. Every generator and every baseline builds its prompt list identically -- the flat annotation list from \texttt{captions\_val2014.json} shuffled with seed 42 -- and writes image $i$ to \texttt{\{i:05d\}.png}, with the latent seed fixed at the global prompt index. A given index therefore selects the same caption, the same latent initialization, and the same scored caption under every configuration, so all comparisons in Table~\ref{tab:combined_results} and Appendix~\ref{app:ablations} are like-for-like.

\paragraph{Reproducing the main table:} Each row of Table~\ref{tab:combined_results} is one generation call followed by one scoring call. For the Chameleon rows, the calibration artifacts are loaded rather than recomputed, for example
  \begin{verbatim}
  python chameleon_generation.py --weight-bits 4 \
      --num-samples 24576 --num-inference-steps 50 --batch-size 16 \
      --coco-captions captions_val2014.json \
      --load-weights  sdxl_w4a8_weights.pt \
      --load-activation-lut sdxl_w4a8_lut.pt \
      --output-dir <out>
  \end{verbatim}
with the per-backbone step counts, guidance scales and resolutions of Appendix~\ref{app:setup}. Baselines are invoked through their own generators at the settings their authors specify; we note in particular that \texttt{MixDQ}'s W$_{4}$A$_{8}$ configuration is a per-layer mixed-precision assignment averaging four bits, supplied as a YAML configuration, rather than uniform 4-bit quantization. Every ablation arm in Appendix~\ref{app:ablations} is a single entry in the released dispatcher, which applies one monkey-patch to the shipped code path and reruns generation and scoring unchanged.

\paragraph{Calibration:} Calibration uses 128 prompts, $B = 10$ timestep buckets, a 20-step trajectory per prompt, and no gradient updates anywhere. The procedure is given as pseudocode in Appendix~\ref{app:algorithms} and its end-to-end cost in Table~\ref{tab:calib_cost}: under an hour for all three backbones combined.

\paragraph{Scoring:} We report clean-\texttt{FID} \citep{parmar2022cleanfid} computed on the full 24{,}576-image generation against the complete 40{,}504-image \texttt{val2014} reference set, and \texttt{CLIP}-score with \texttt{ViT-L/14} \citep{radford2021clip} against the caption each image was generated from; the scorer reconstructs the generators' shuffled caption list verbatim (Appendix~\ref{app:clip_scaling}). Sharpness measurements in Appendix~\ref{app:qualitative} use the variance of the $3 \times 3$ discrete Laplacian of the grayscale image.

\paragraph{Known deviations:} Two properties of the implementation should be noted by anyone reproducing or extending this work. All \rev{quality} results use fake-quantization -- round-trip casts inside \texttt{FP16} -- rather than low-precision kernels\rev{; real-kernel speed and fidelity are reported in Appendix~\ref{app:rebuttal}}. And on \texttt{SDXL} our weight-side \texttt{MX} implementation uses an unconstrained per-block scale rather than the power-of-two \texttt{E8M0} exponent the \texttt{OCP} specification requires (the few-step path already conforms, and the \texttt{DiT} path selects no \texttt{MX} weight format); Appendix~\ref{app:mxconformance} quantifies what conformance would cost.

\subsection{Acknowledgments}
\label{stat:ack}

The authors acknowledge the Texas Advanced Computing Center (TACC) at The University of Texas at Austin for providing computational resources that have contributed to the research results reported within this paper. URL: \url{http://www.tacc.utexas.edu}

\end{document}